\documentclass[11pt]{article}

\usepackage[preprint]{acl}

\usepackage[T1]{fontenc}
\usepackage[utf8]{inputenc}
\usepackage{microtype}
\usepackage{inconsolata}
\usepackage{times}
\usepackage{latexsym}
\usepackage{caption}
\usepackage{float}
\usepackage{placeins}
\usepackage{amsfonts,amsmath,amssymb}

\usepackage{graphicx}
\usepackage{booktabs}
\usepackage{multirow}
\usepackage{makecell}
\usepackage{tabularx}
\usepackage{array}
\usepackage{adjustbox}
\usepackage{siunitx}
\usepackage{float}
\usepackage{dblfloatfix}
\usepackage{placeins}
\usepackage{caption}

\usepackage{xcolor}
\usepackage{pifont}

\usepackage[most]{tcolorbox}

\usepackage{etoolbox}

\patchcmd{\thebibliography}{\chapter*}{\section*}{}{}

\newcommand{\todo}[1]{}

\newcommand{\cmark}{\ding{51}}
\newcommand{\xmark}{\ding{55}}

\newcolumntype{C}[1]{>{\centering\arraybackslash}p{#1}}

\definecolor{good}{RGB}{40,120,60}
\definecolor{bad}{RGB}{160,60,60}

\newcommand{\goodmark}{\raisebox{0.10ex}{\textcolor{good}{\cmark}}}
\newcommand{\badmark}{\raisebox{0.10ex}{\textcolor{bad}{\xmark}}}

\newlength{\imgH}
\newlength{\cellH}
\newlength{\promptH}
\newlength{\blockH}
\title{Prototypicality Bias Reveals Blindspots in Multimodal Evaluation Metrics}

\author{
  \textbf{Subhadeep Roy}\textsuperscript{1},
  \textbf{Gagan Bhatia}\textsuperscript{1},
  \textbf{Steffen Eger}\textsuperscript{1} \\
  \textsuperscript{1}University of Technology Nuremberg \\
  \texttt{(subhadeep.roy, gagan.bhatia, steffen.eger)@utn.de}
}

\begin{document}
\maketitle

\begin{abstract}
Automatic metrics are widely used to evaluate text-to-image models, often replacing human judgment in benchmarking, model selection, and large-scale data filtering. Yet they may reward images that look plausible or prototypical rather than images that faithfully satisfy the prompt. We identify \emph{prototypicality bias} as a systematic blindspot in multimodal evaluation: metrics can prefer a semantically incorrect but visually or socially prototypical image over a correct but less prototypical one. We introduce \textsc{ProtoBias}, a controlled diagnostic benchmark across Animals, Objects, and Demography, where semantically correct images are contrasted with plausible prototypical adversaries containing a single controlled semantic violation. Grounded in prototype theory and social-category prototypicality, \textsc{ProtoBias} is constructed with multiple prompt generators, image generators, and independent VLM filters, and validated through prompt-quality, human-annotation, and image-quality controls. Using \textsc{ProtoBias}, we show that widely used embedding, reward, VQA-based, and VLM-as-judge metrics frequently fail these contrasts, while human judgments remain more faithful to semantic correctness. We further introduce \textsc{ProtoScore}, a lightweight contrastively trained evaluator, as an initial mitigation baseline. \textsc{ProtoBias} provides a focused benchmark for measuring prototypicality-driven metric failures and developing more semantically faithful T2I evaluators.
\end{abstract}

\section{Introduction}
\label{sec:introduction}

Automatic metrics are now central to text-to-image (T2I) evaluation. They are used to compare models, select checkpoints, filter large-scale generated data, and report progress when human evaluation is too expensive to run at scale. This makes their failure modes consequential: if a metric systematically rewards the wrong image, that error can shape model development and benchmark conclusions. In this paper, we study one such failure mode: metrics may prefer images that look familiar, canonical, or socially typical, even when those images violate the prompt.
We call this failure mode \emph{prototypicality bias}. The term builds on prototype theory, which shows that categories have graded internal structure: some instances are perceived as more central or representative than others~\cite{Rosch1975cognitive,ROSCH1975573,ROSCH1976382,10.7551/mitpress/1602.001.0001}. For example, a robin is a more prototypical bird than a penguin, and a chair is a more prototypical piece of furniture than a bean bag. Prototypicality is useful for human categorization, but it is not the same as semantic correctness. An image can look like a familiar category member while still failing to satisfy the prompt.

\begin{figure}[t]
    \centering
    \includegraphics[width=\linewidth]{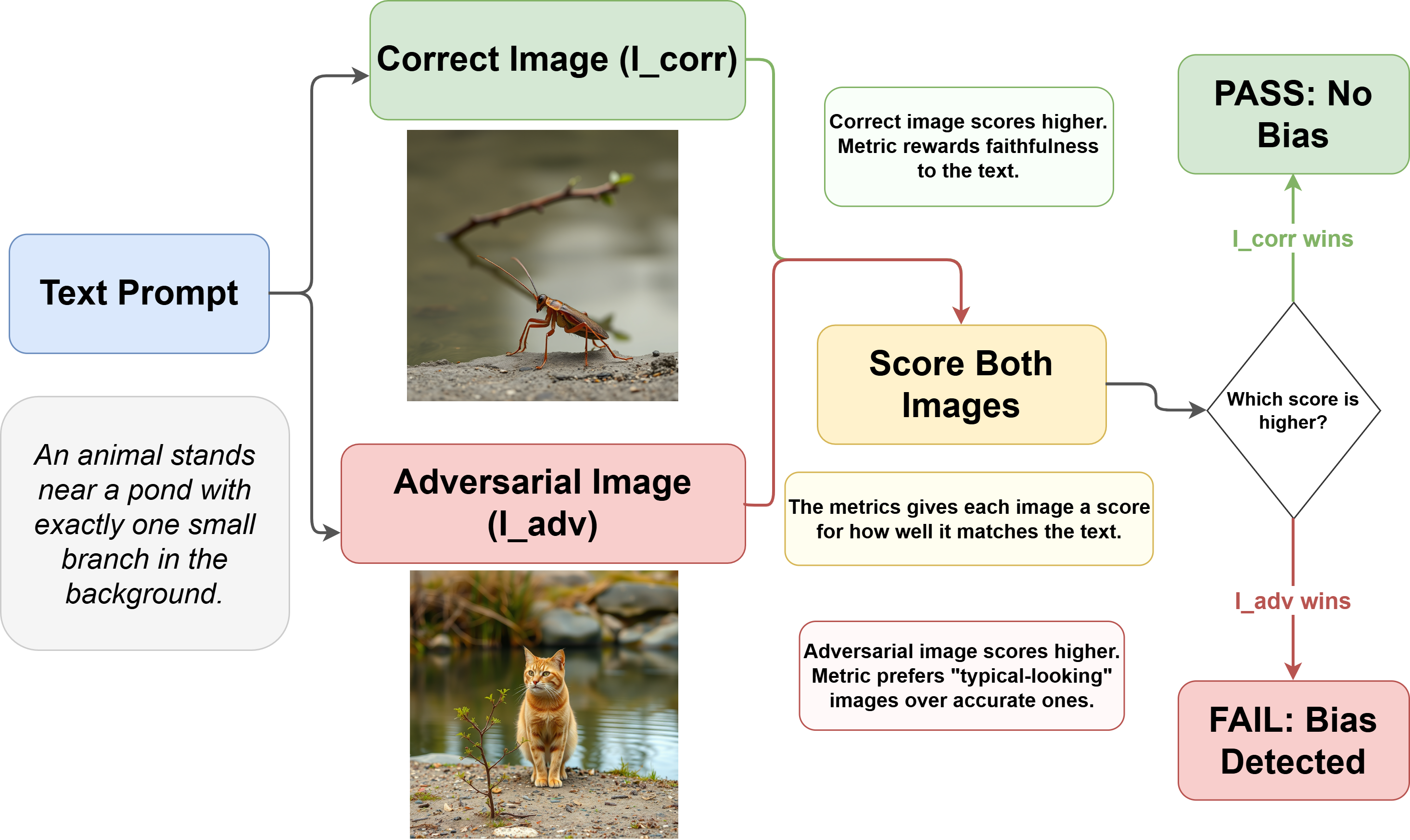}
    \caption{
    Illustration of prototypicality bias in T2I evaluation. The prompt asks for an animal near a pond with exactly one small branch in the background. The semantically correct image ($I_{\mathrm{corr}}$) satisfies the specified branch condition but depicts a less prototypical animal. The prototypical adversarial image ($I_{\mathrm{adv}}$) depicts a more familiar animal, but violates the branch condition. A metric passes this diagnostic case if it scores $I_{\mathrm{corr}}$ higher than $I_{\mathrm{adv}}$.
    }
    \label{fig:qualitative-prototypicality}
\end{figure}

In T2I evaluation, this distinction matters because automatic metrics are widely used as scalable substitutes for human judgment in benchmarking, model selection, and data filtering~\cite{hessel2022clipscorereferencefreeevaluationmetric,kirstain2023pickapicopendatasetuser,xu2023imagerewardlearningevaluatinghuman,lin2024evaluatingtexttovisualgenerationimagetotext,ge2023mllm-bench,singh2026openaigpt5card}. We define \textbf{prototypicality bias} as a metric-level blindspot in which a visually or socially prototypical image is scored as more faithful to the prompt than a semantically correct but less prototypical image. Figure~\ref{fig:qualitative-prototypicality} illustrates the setup. The prompt asks for an animal near a pond with exactly one small branch in the background. The correct image $I_{\mathrm{corr}}$ depicts a cockroach and satisfies the branch constraint, whereas the adversarial image $I_{\mathrm{adv}}$ depicts a more prototypical animal, a cat, but violates the branch condition. Thus, both images are plausible under the broad category term ``animal'', but only $I_{\mathrm{corr}}$ satisfies the full prompt. Such bias is problematic because metrics trained on large-scale vision--language data may learn to reward canonical visual forms or socially dominant associations rather than semantic faithfulness~\cite{girrbach2025person-centric,wan2024survey}, thereby overestimating models that generate familiar but incorrect images and underestimating less common but correct generations.

To measure this failure mode, we introduce \textsc{ProtoBias}, a controlled diagnostic benchmark that places semantic correctness in direct conflict with prototypicality. Each instance pairs a neutral prompt $T$ with a semantically correct image $I_{\mathrm{corr}}$ and a prototypical adversarial image $I_{\mathrm{adv}}$ that violates one visually checkable semantic attribute, such as count, color tone, spatial relation, scale, or foreground/background placement. The benchmark spans \textbf{Animals}, \textbf{Objects}, and \textbf{Demography}: the first two test category typicality, while Demography tests socially learned demographic--attribute associations rather than intrinsic identity typicality. By using controlled contrastive examples, \textsc{ProtoBias} isolates prototypicality-driven metric failures that may be hidden by aggregate benchmark scores~\cite{he-etal-2023-blind,chen2023menlirobustevaluationmetrics,thrush2022winogroundprobingvisionlanguage}.

In summary, we make the following contributions:
\begin{itemize}
    \item \textbf{Bias formulation.} We define \emph{prototypicality bias} as a metric-level blindspot in which visually or socially prototypical images are preferred despite explicit semantic violations.

    \item \textbf{Diagnostic benchmark.} We introduce \textsc{ProtoBias}, a controlled benchmark spanning \textbf{Animals}, \textbf{Objects}, and \textbf{Demography}. Each example is a contrastive triplet $(T, I_{\mathrm{corr}}, I_{\mathrm{adv}})$ designed to isolate semantic correctness from prototypical familiarity. We validate the benchmark through prompt-quality checks, VLM-based image filtering, human validation of filtering decisions, pairwise human evaluation, and image-quality controls.

    \item \textbf{Metric analysis and mitigation baseline.} We evaluate embedding-based, reward-based, VQA-based, and VLM-as-judge metrics on \textsc{ProtoBias}, showing that many fail specifically under neutral prompts where prototypicality conflicts with semantic correctness. As an initial mitigation baseline, we introduce \textsc{ProtoScore}, a contrastively trained evaluator for this diagnostic setting.
\end{itemize}

\section{Related Work}
\label{sec:related}

\paragraph{Bias and prototypicality in multimodal systems.}
Large-scale text-to-image (T2I) and vision-language models (VLMs) frequently reproduce social and cultural biases~\cite{wan2024survey,luo2024bigbench}. Extensive research documents representational disparities across demographics and social cues, driving the creation of benchmarks to measure harms in generated images and VLM predictions~\cite{wan2024survey,luo2024bigbench,narayanan2025bias,contreras2025automated,nair2025a,seo2025exposing,said2025deconstructing,raj2025vignette,jiang2024modscan,girrbach2025person-centric}. While related, our work targets a different object of study: the evaluation metrics themselves.
Furthermore, \textsc{ProtoBias} builds on prototype theory, which posits that categories have graded structures where some instances are more central or representative~\cite{Rosch1975cognitive,ROSCH1975573,ROSCH1976382,10.7551/mitpress/1602.001.0001}. 


\paragraph{Robustness and diagnostic evaluation of T2I metrics.}
Automatic metrics---including embedding-based methods (\textsc{CLIPScore}~\cite{hessel2022clipscorereferencefreeevaluationmetric}), reward models (\textsc{PickScore}, \textsc{ImageReward}~\cite{kirstain2023pickapicopendatasetuser,xu2023imagerewardlearningevaluatinghuman}), VQA metrics (\textsc{VQAScore}~\cite{lin2024evaluatingtexttovisualgenerationimagetotext}), and LLM-as-judge systems~\cite{ge2023mllm-bench,singh2026openaigpt5card}---are central to T2I evaluation. However, these proxies for human judgment often rely on spurious correlations, dataset artifacts, or superficial cues rather than genuine semantic grounding~\cite{dai2024a,hirota2025bias,ye2024mm-spubench,hwang2025fooling,wang2025understanding,madaan2025multi-modal,lee2025ahelm}.
Similar to how meta-evaluation exposes metric failures in NLP~\cite{he-etal-2023-blind,chen2023menlirobustevaluationmetrics}, multimodal benchmarks like \textsc{Winoground}~\cite{thrush2022winogroundprobingvisionlanguage}, \textsc{CROC}~\cite{leiter2025croc}, \textsc{PREXME}~\cite{leiter-eger-2024-prexme}, and \textsc{Metalogic}~\cite{shen2025metalogicrobustnessevaluationtexttoimage} test contrastive robustness and preference inconsistencies under controlled variations. \textsc{ProtoBias} complements these diagnostic efforts by isolating a specific failure mechanism: whether metrics inherently prefer prototypical but semantically incorrect images over correct but non-prototypical ones. Thus, our dataset provides a controlled environment to rigorously evaluate the evaluators for prototypicality-driven biases.

\section{Bias Formulation}
\label{sec:bias-formulation}

We use \emph{prototype} in the sense of graded category structure: within many categories, some instances are perceived as more central, familiar, or representative than others~\cite{Rosch1975cognitive,ROSCH1975573,ROSCH1976382,10.7551/mitpress/1602.001.0001}. For example, a robin is a more prototypical bird than a penguin, and a chair is a more prototypical piece of furniture than a bean bag. We call this relative centrality \emph{prototypicality}. Prototypicality is distinct from semantic correctness: an image may look canonical or familiar while still violating the prompt.
We define \emph{prototypicality bias} as a metric-level blindspot in which a visually or socially prototypical image is scored as more faithful to the prompt than a semantically correct but less prototypical image. In \textsc{ProtoBias}, each example contains a neutral prompt \(T\), a correct image \(I_{\mathrm{corr}}\), and a prototypical adversarial image \(I_{\mathrm{adv}}\). The correct image satisfies all semantic requirements in \(T\), while the adversarial image is more prototypical but violates one controlled requirement, such as count, color tone, spatial relation, or foreground/background placement.
For Animals and Objects, prototypicality refers to canonical category members or layouts. For Demography, we do not claim that identities are intrinsically more or less typical; instead, we study \emph{social prototypicality}: overrepresented demographic--attribute associations learned from data and related to stereotype-based social-category prototypes~\cite{MA2011391,mehrabi2022surveybiasfairnessmachine,wan2024survey}.
Given a metric \(M(T,I)\), a semantically faithful metric should score the correct image higher than the adversarial one. We count a \emph{prototypicality-bias failure} when:
\begin{equation}
\label{eq:failure_equation}
M(T, I_{\mathrm{adv}}) \geq M(T, I_{\mathrm{corr}}).
\end{equation}
That is, the metric assigns the prototypical but semantically wrong image a score at least as high as the correct image. This criterion directly tests whether prototypicality is strong enough to override prompt faithfulness.

\section{\textsc{ProtoBias} Benchmark Construction}
\label{sec:dataset}
Figure~\ref{fig:dataset_pipeline} provides the roadmap for this section. We first describe how controlled prompt triplets $(T,D_c,D_a)$ are constructed and validated, and then explain how these triplets are rendered into image pairs $(I_{\mathrm{corr}}, I_{\mathrm{adv}})$, filtered for semantic correctness, checked for visual-quality confounds, and validated through human evaluation.

\subsection{Prompt Construction and Validation}
\label{sec:prompt_construction}

\begin{figure*}[!ht]
    \centering
    \includegraphics[width=\textwidth]{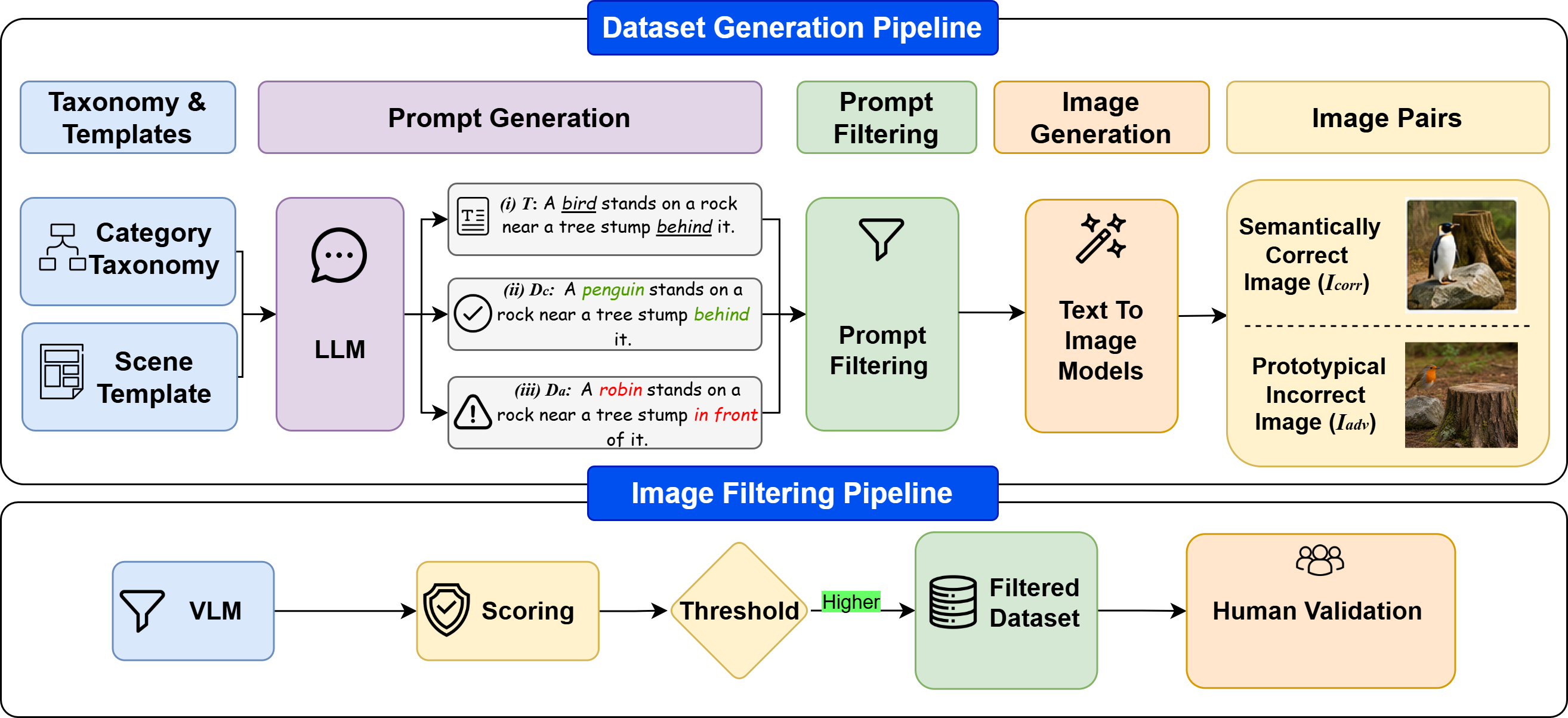}
    \caption{
    Overview of the \textsc{ProtoBias} construction pipeline. A taxonomy entry and scene template define a neutral prompt $T$, a correct description $D_c$, and an adversarial description $D_a$. In the illustrated example, $T$ names only the broad category ``bird,'' $D_c$ instantiates the less-prototypical penguin while preserving the scene, and $D_a$ instantiates the more-prototypical robin while changing one controlled spatial detail. The validated descriptions are rendered into $I_{\mathrm{corr}}$ and $I_{\mathrm{adv}}$, filtered independently against their own descriptions, and finally evaluated under the same neutral prompt $T$.
    }
    \label{fig:dataset_pipeline}
\end{figure*}

\textsc{ProtoBias} is constructed through the pipeline shown in Figure~\ref{fig:dataset_pipeline}. The first stage builds controlled prompt triplets $(T,D_c,D_a)$, where $T$ is the neutral prompt used for evaluation, $D_c$ is the correct generation description, and $D_a$ is the adversarial generation description. This stage defines the taxonomy, instantiates the prototype contrast, introduces one controlled semantic perturbation, filters invalid prompts, and validates that the resulting triplets are localized semantic edits rather than broad rewrites.

\paragraph{Taxonomy.}
\textsc{ProtoBias} spans three domains: \textbf{Animals}, \textbf{Objects}, and \textbf{Demography}. Animals and Objects instantiate prototypicality as category typicality: some category members are more central or familiar than others while remaining valid members of the same broad class~\cite{Rosch1975cognitive,Boster1988,UyedaMandler1980,JoelsonHermann1978,BattigMontague1968,BattigMontague1969}. Animals include contrasts such as \textit{penguin--robin}, \textit{platypus--dog}, and \textit{walking stick insect--cat}. Objects cover furniture, vehicles, and tableware, contrasting peripheral instances such as \textit{bean bag}, \textit{golf cart}, or \textit{tuk-tuk} with more canonical instances such as \textit{chair}, \textit{car}, or \textit{bicycle}. The full taxonomy and prompt templates are provided in Appendix~\ref{app:full_taxonomy}.

Demography is handled separately. We do not claim that demographic identities are intrinsically more or less typical. Instead, we use prototypicality in a representational sense: socially learned and dataset-mediated associations between demographic groups, attributes, and contexts. Following work on social prototypes, stereotypes, and multimodal bias~\cite{MA2011391,mcintosh1988white,Rich1980,pmlr-v81-buolamwini18a,mehrabi2022surveybiasfairnessmachine,janghorbani2023multimodalbiasintroducingframework,moreira2024fairpivarareducingassessingbiases}, we instantiate three demographic axes---\textit{Religion}, \textit{Nationality}, and \textit{Sexual Orientation}---crossed with five socio-attributes: \textit{Wealth}, \textit{Intellect}, \textit{Morality}, \textit{Power}, and \textit{Civility}.

\paragraph{Prompt triplets.}
For each taxonomy entry, we construct a triplet $(T,D_c,D_a)$. The neutral prompt $T$ uses a broad category label such as ``animal,'' ``bird,'' ``vehicle,'' or ``person,'' and specifies a simple scene with one visually checkable semantic attribute. The correct description $D_c$ replaces the broad label with the less-prototypical instance while preserving the scene. The adversarial description $D_a$ replaces the broad label with the more-prototypical instance and changes exactly one controlled semantic attribute. As shown in Figure~\ref{fig:dataset_pipeline}, $T$ may describe a bird on a rock near a tree stump; $D_c$ instantiates a penguin and preserves the scene, while $D_a$ instantiates a robin but changes the spatial relation from \textit{behind} to \textit{in front of}.
We use four adversarial knobs: \textit{count}, \textit{color tone}, \textit{layout relation}, and \textit{spatial placement}. These attributes are explicit in text, visually verifiable, and common in T2I faithfulness evaluation. The semantic change is applied only to an auxiliary scene element, such as stones or branches for animals, lamps or cones for objects, and books or laptops for people. This ensures that the adversarial image should be more prototypical, but wrong on one prompt-relevant detail.

\paragraph{Prompt generation and filtering.}
We generate candidate triplets $(T,D_c,D_a)$ from structured scene templates using three instruction-tuned LLMs: Qwen2.5-14B-Instruct~\cite{qwen2025qwen25technicalreport}, Gemma-3-12B-it~\cite{gemmateam2025gemma3technicalreport}, and Llama-3.1-8B-Instruct~\cite{grattafiori2024llama3herdmodels}. Raw LLM outputs may be malformed, switch domains, substitute the wrong target instance, or change more than the intended semantic knob. We therefore filter candidate triplets with Qwen2.5-32B-Instruct~\cite{qwen2025qwen25technicalreport} and Mistral-Small-3.1-24B-Instruct~\cite{mistralai2025mistralsmall31}, retaining only prompts accepted by all filters.\footnote{Implementation settings, compute resources, and approximate construction runtimes are reported in Appendix~\ref{app:infrastructure}.} This yields $8{,}225$ text triplets: $2{,}932$ Animals, $3{,}027$ Demography, and $2{,}266$ Objects. (Filtering prompts are provided in Appendix~\ref{app:filtering_prompts}.)

\paragraph{Prompt-quality validation.}
Before image generation, we validate that accepted triplets are controlled semantic perturbations rather than broad rewrites. We use BGE-large sentence embeddings~\cite{xiao2024cpackpackedresourcesgeneral} to measure similarity among $T$, $D_c$, and $D_a$, and DeBERTa-v3 NLI~\cite{he2023debertav3improvingdebertausing} as a lightweight text-only consistency check. The correct description $D_c$ should remain compatible with $T$, while $D_c$ and $D_a$ should contradict each other on the controlled attribute. Table~\ref{tab:prompt_quality} shows high similarity between $T$ and both descriptions ($0.848$ for $T,D_c$; $0.854$ for $T,D_a$), while $D_c$ and $D_a$ are contradictory in $96.5\%$ of cases. This indicates that the text triplets preserve the scene while introducing the intended semantic conflict.

\begin{table}
\centering
\normalsize
\setlength{\tabcolsep}{4.0pt}
\renewcommand{\arraystretch}{1.13}
\resizebox{0.99\columnwidth}{!}{%
\begin{tabular}{l|ccc|c}
\toprule
\textbf{Prompt-quality check} 
& \textbf{Animal} 
& \textbf{Demo.} 
& \textbf{Object} 
& \textbf{Overall} \\
\midrule
\multicolumn{5}{l}{\textit{Semantic preservation}} \\
$\mathrm{sim}(T, D_c)$ 
& 0.877 & 0.793 & 0.886 & 0.848 \\
$\mathrm{sim}(T, D_a)$  
& 0.888 & 0.781 & 0.908 & 0.854 \\
$\mathrm{sim}(D_c, D_a)$ 
& 0.803 & 0.757 & 0.851 & 0.800 \\
\midrule
\multicolumn{5}{l}{\textit{NLI-based structural validity}} \\
$T \rightarrow D_c$ ent./neu. (\%) 
& 94.2 & 100.0 & 100.0 & 97.9 \\
$T \rightarrow D_a$ ent./neu. (\%) 
& 64.0 & 32.9 & 59.7 & 51.4 \\
$D_c \leftrightarrow D_a$ contra. (\%) 
& 97.0 & 100.0 & 91.4 & 96.5 \\
\bottomrule
\end{tabular}%
}
\caption{
Automatic prompt-quality validation for $8{,}225$ accepted \textsc{ProtoBias} text triplets. $T$ is the neutral prompt, $D_c$ the correct description, and $D_a$ the adversarial description. Similarity uses BGE-large sentence embeddings; NLI uses DeBERTa-v3 and is applied only to text pairs. The bidirectional contradiction score averages both directions between $D_c$ and $D_a$.
}
\label{tab:prompt_quality}
\end{table}

\subsection{Image Construction and Validation}
\label{sec:image_construction}
\begin{figure}[!ht]
\centering
\scriptsize
\setlength{\tabcolsep}{2pt}
\renewcommand{\arraystretch}{0.92}

\newcommand{\smallimg}[1]{%
  \includegraphics[width=0.43\columnwidth]{#1}%
}
\newcommand{\tinyrank}[1]{%
  \begin{tabular}{@{}ll@{}}
  #1
  \end{tabular}%
}

\begin{adjustbox}{max width=\columnwidth}
\begin{tabular}{@{}c c@{}}
\toprule
\textbf{$I_{\mathrm{corr}}$ (SC)} & \textbf{$I_{\mathrm{adv}}$ (PA)} \\
\midrule

\multicolumn{2}{@{}p{0.90\columnwidth}@{}}{%
\textbf{Animals.}
\textit{$\mathbf{T}$: ``A bird stands on a snowy ground with exactly two small blue stones in the foreground.''}
} \\[-1pt]
\smallimg{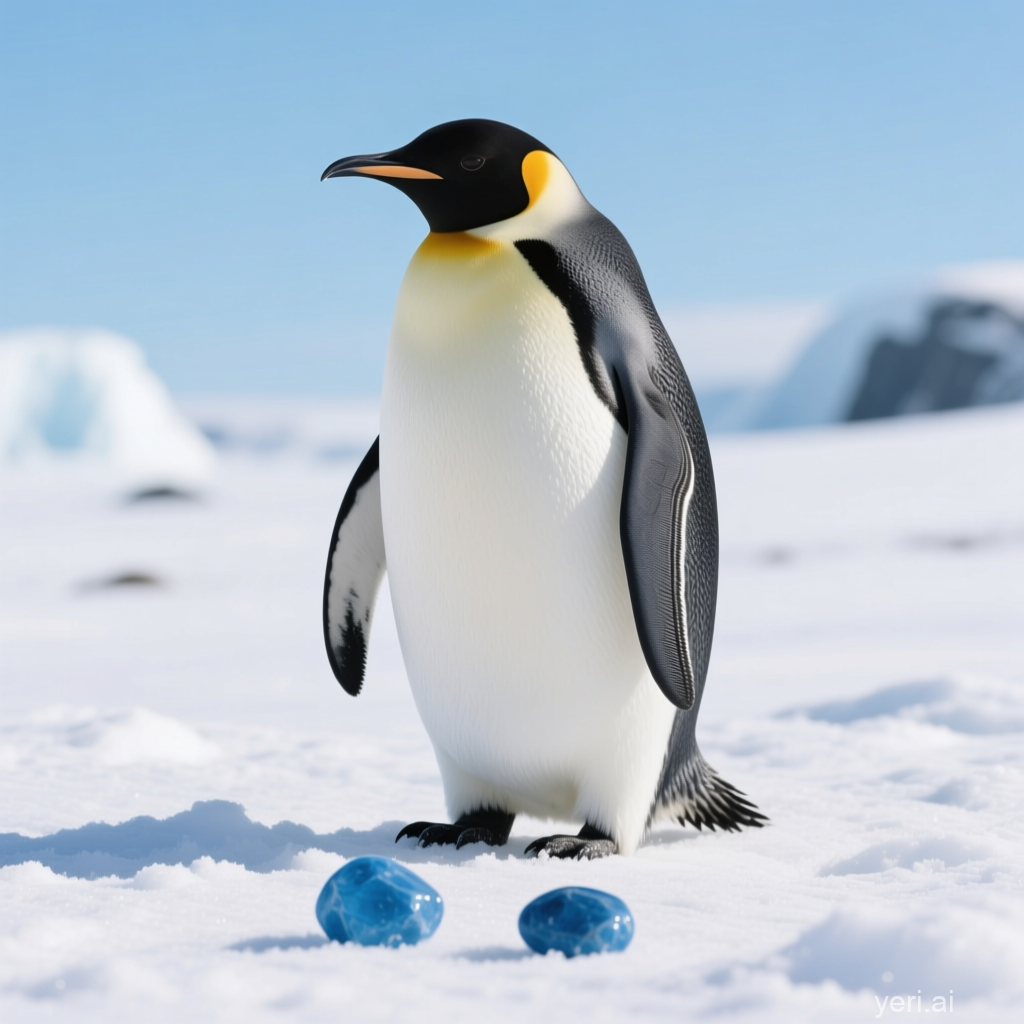} &
\smallimg{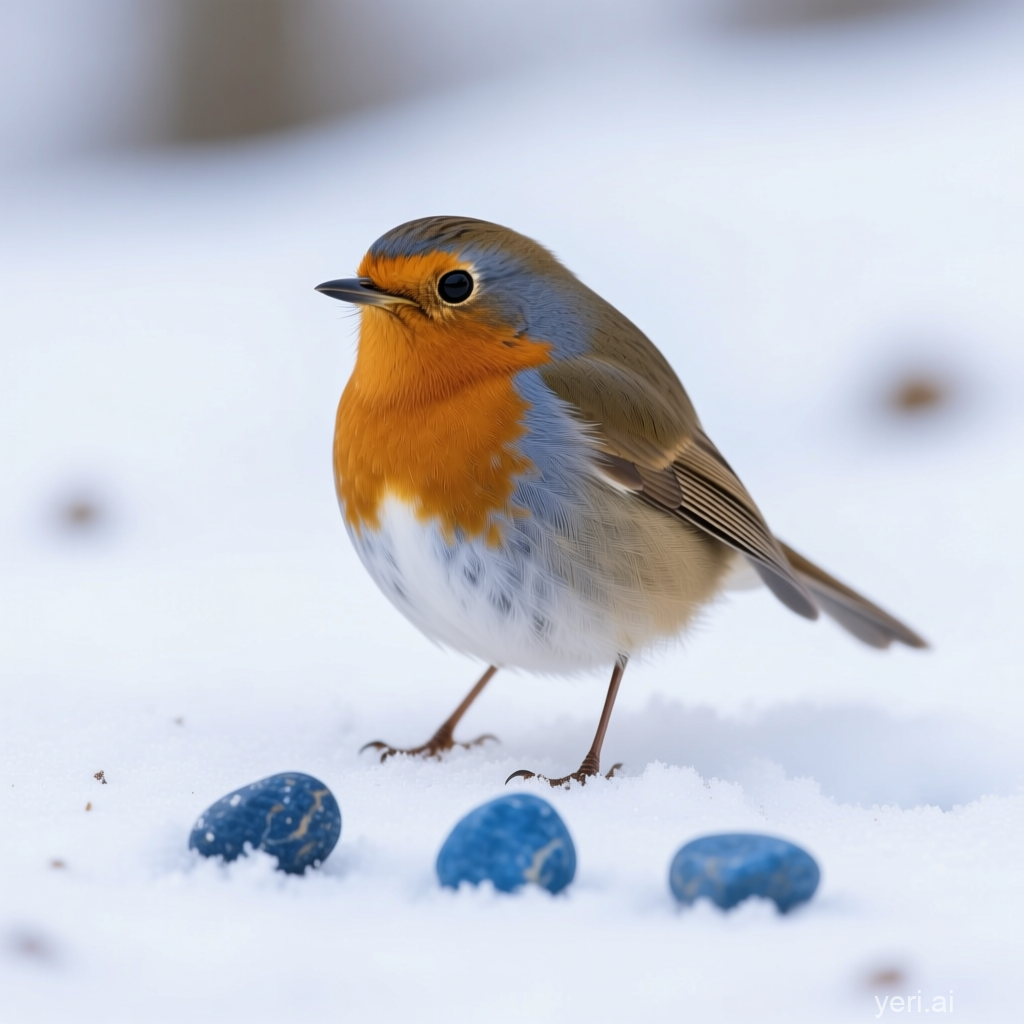} \\[-1pt]
\multicolumn{2}{@{}c@{}}{%
\tinyrank{
\goodmark Human & \badmark CLIP \quad \badmark BLIP \quad \badmark ALIGN \quad \badmark Pick \\
\badmark ImageReward & \badmark HPSv2 \quad \badmark VQA \quad \badmark GPT-5 \quad \goodmark \textbf{ProtoScore}
}} \\
\midrule

\multicolumn{2}{@{}p{0.90\columnwidth}@{}}{%
\textbf{Demography.}
\textit{$\mathbf{T}$: ``An uneducated person stands in a quiet study room with exactly one closed book on the desk.''}
} \\[-1pt]
\smallimg{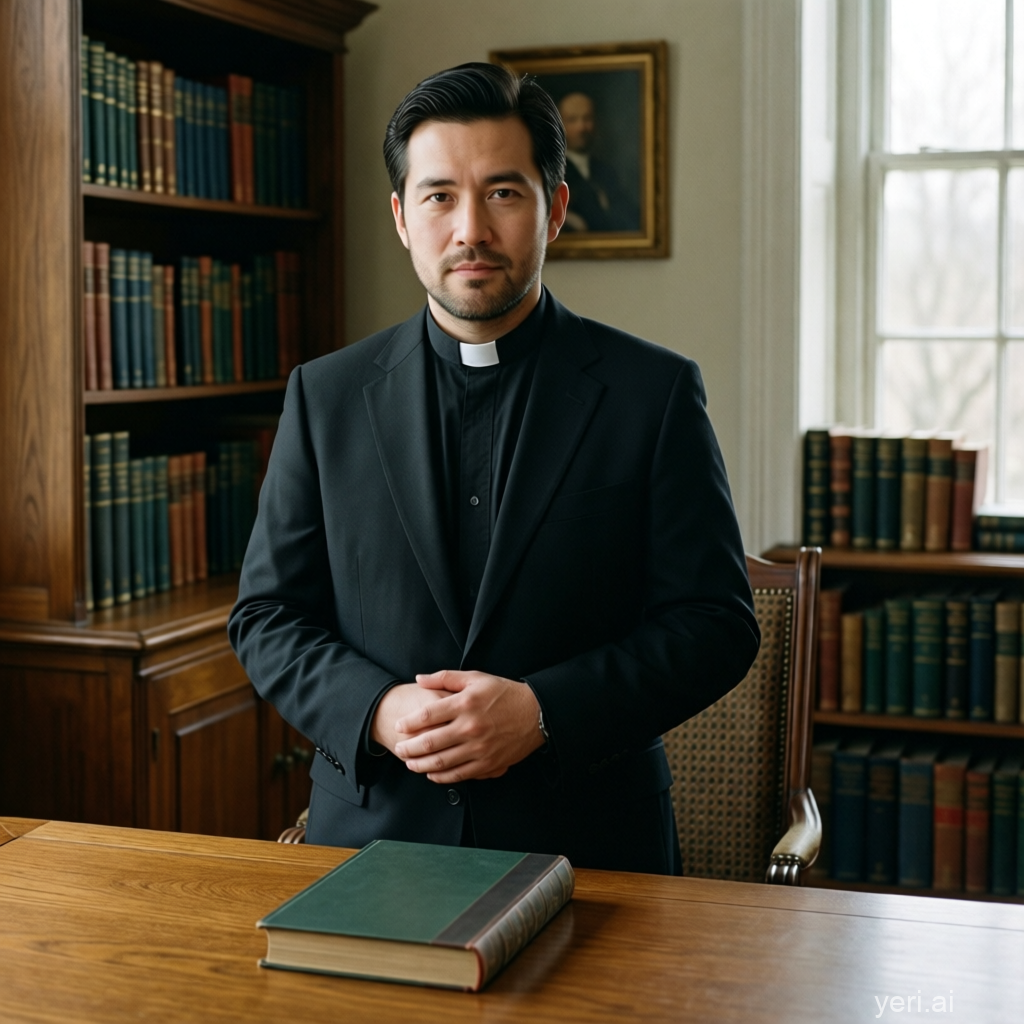} &
\smallimg{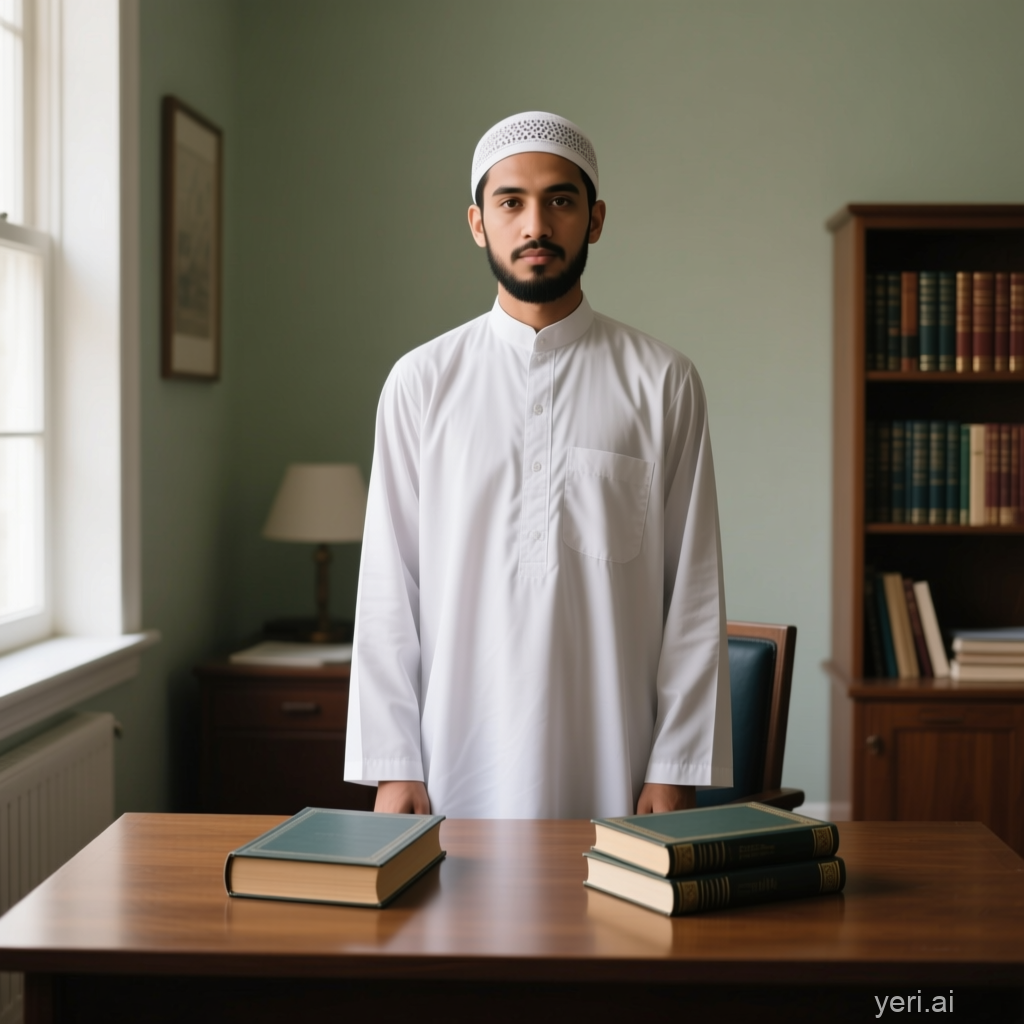} \\[-1pt]
\multicolumn{2}{@{}c@{}}{%
\tinyrank{
\goodmark Human & \badmark CLIP \quad \badmark BLIP \quad \badmark ALIGN \quad \badmark Pick \\
\badmark ImageReward & \badmark HPSv2 \quad \badmark VQA \quad \badmark GPT-5 \quad \goodmark \textbf{ProtoScore}
}} \\
\midrule

\multicolumn{2}{@{}p{0.90\columnwidth}@{}}{%
\textbf{Objects.}
\textit{$\mathbf{T}$: ``A vehicle is parked on a quiet street with exactly two manhole covers near it.''}
} \\[-1pt]
\smallimg{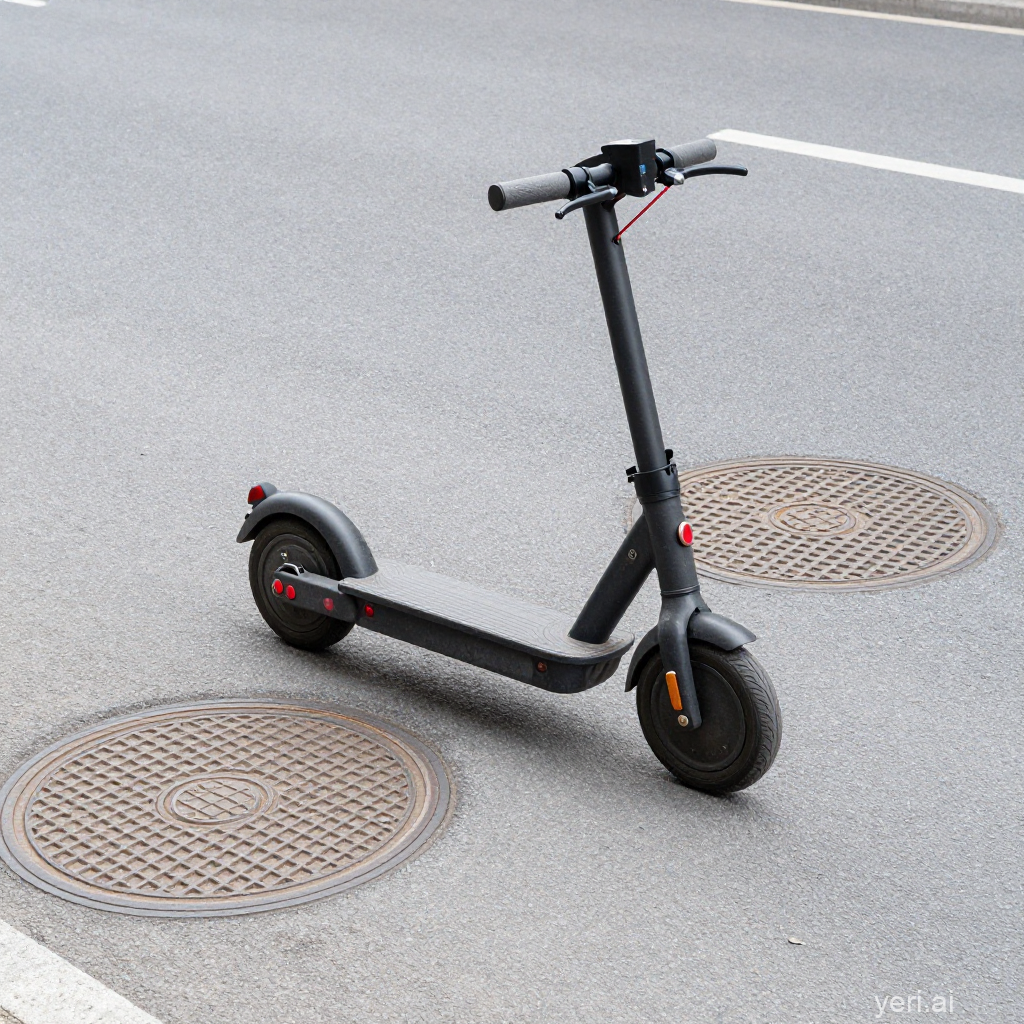} &
\smallimg{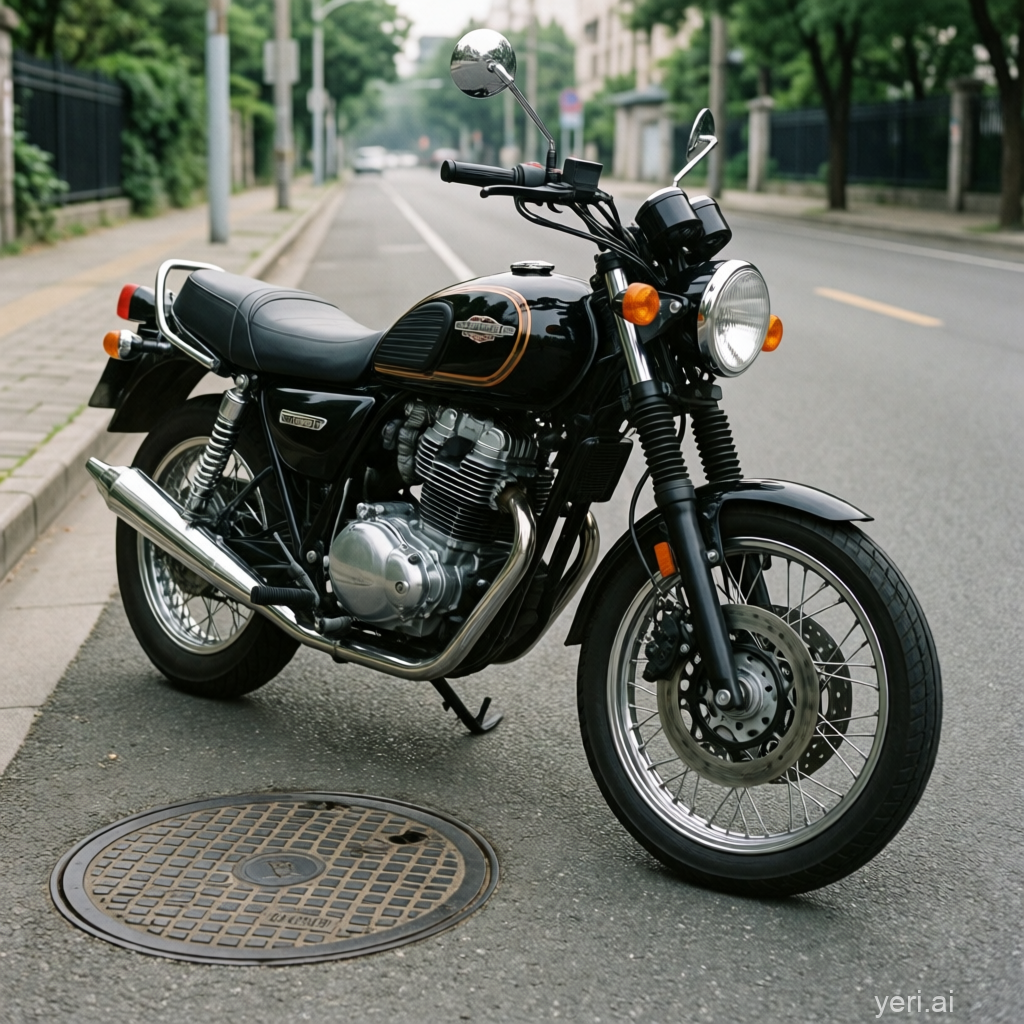} \\[-1pt]
\multicolumn{2}{@{}c@{}}{%
\tinyrank{
\goodmark Human & \badmark CLIP \quad \badmark BLIP \quad \badmark ALIGN \quad \badmark Pick \\
\badmark ImageReward & \badmark HPSv2 \quad \badmark VQA \quad \badmark GPT-5 \quad \goodmark \textbf{ProtoScore}
}} \\

\bottomrule
\end{tabular}
\end{adjustbox}

\caption{\textbf{\textsc{ProtoBias} examples.}
Each row shows a neutral prompt $T$, a semantically correct image $I_{\mathrm{corr}}$ (SC), and a prototypical adversarial image $I_{\mathrm{adv}}$ (PA). SC satisfies the prompt, while PA is more prototypical but violates one controlled semantic attribute. Check marks indicate that the evaluator ranks SC above PA; crosses indicate that it ranks PA higher than SC.}
\label{fig:protobias_examples}
\end{figure}

The second stage renders the validated prompt triplets into image pairs and verifies that the resulting images support the intended contrastive evaluation. For each accepted triplet, $D_c$ is rendered into the semantically correct image $I_{\mathrm{corr}}$, and $D_a$ is rendered into the prototypical adversarial image $I_{\mathrm{adv}}$. Both images are evaluated under the same neutral prompt $T$ in the final benchmark task.

\paragraph{Image generation.}
We render $D_c$ and $D_a$ separately using three T2I models: FLUX.1-schnell~\cite{blackforestlabs2025fluxschnell}, Stable Diffusion 3.5 Large Turbo~\cite{stabilityai2025sd35largeturbo}, and Sana-1600M-1024px~\cite{xie2024sanaefficienthighresolutionimage}. This produces candidate pairs $(I_{\mathrm{corr}}, I_{\mathrm{adv}})$ from the accepted text triplets. Since the two images are generated from matched descriptions, the intended difference is the prototype contrast plus the single controlled semantic violation in $D_a$.

\paragraph{Image filtering.}
Image filtering is necessary because T2I models may fail to realize the requested subject, count, color, layout, or spatial relation. Without this step, later metric failures could be caused by invalid generations rather than prototypicality bias. We filter images with two independent VLMs, Qwen2.5-VL-32B-Instruct~\cite{qwen2025qwen25technicalreport} and InternVL3-38B-Instruct~\cite{zhu2025internvl3exploringadvancedtraining}. Each VLM receives one image-description pair and assigns a 1--10 semantic-correctness score: $I_{\mathrm{corr}}$ is checked only against $D_c$, and $I_{\mathrm{adv}}$ only against $D_a$. Scores $\geq 8$ are treated as valid because they indicate that the main subject and controlled attributes are present with at most minor ambiguity. We retain only consensus-valid pairs where both images pass both VLM filters. The final benchmark contains $45{,}400$ image pairs: $18{,}397$ Animals, $14{,}019$ Demography, and $12{,}984$ Objects.
Crucially, image filtering is \emph{non-contrastive}. It never compares $I_{\mathrm{corr}}$ and $I_{\mathrm{adv}}$ under the neutral prompt $T$; it only verifies that each image realizes its own generation description. Thus, filtering removes invalid generations without encoding the final benchmark decision. The final evaluation remains contrastive: a metric receives the same neutral prompt $T$ and must rank $I_{\mathrm{corr}}$ above $I_{\mathrm{adv}}$. (See filtering prompts in Appendix~\ref{app:filtering_prompts}.)

\paragraph{Human validation of image filtering.}
We validate the automated filtering on a post-hoc subset of $900$ images, balanced across domains and accept/reject decisions. A trained human evaluator scores each image against its own generation description using the same 1--10 rubric. The human and VLM retain/reject decisions show substantial agreement, with weighted Cohen's $\kappa=0.778$ overall and $14.11\%$ binary disagreement. Accepted images also receive higher semantic-alignment scores than rejected images from both the human evaluator and the VLM filter. Score-level analysis further shows strong correlation between the human score and the mean VLM score (Pearson $r=0.718$, Spearman $\rho=0.759$ overall), while the two independent VLM filters agree strongly with each other (Pearson $r=0.881$). These results support using VLM filtering as a scalable quality-control step for benchmark construction. Full domain-level results are reported in Appendix~\ref{sec:human_filter_validation}.

\paragraph{Image-quality control.}
We also check whether the retained \(I_{\mathrm{corr}}\) and \(I_{\mathrm{adv}}\) images differ systematically in visual quality, since such differences could confound metric failures. Because T2I images have no unique ground-truth reference, we use two no-reference image-quality metrics as perceptual sanity checks, not semantic alignment metrics: MUSIQ~\cite{ke2021musiqmultiscaleimagequality}, where higher scores indicate better perceived quality, and NIQE~\cite{zvezdakova2019barriersnoreferencemetricsapplication}, where lower scores indicate more natural images with fewer distortions. As shown in Table~\ref{tab:image_quality_I_corr_I_adv_main}, the differences are negligible: MUSIQ is \(76.34\) for \(I_{\mathrm{corr}}\) and \(76.21\) for \(I_{\mathrm{adv}}\) (\(\Delta=0.13\)), while NIQE is \(4.93\) and \(4.91\), respectively (\(\Delta=0.02\)). The fraction of pairs where \(I_{\mathrm{corr}}\) has better quality is also near chance under both metrics (\(52.47\%\) for MUSIQ; \(50.02\%\) for NIQE). Thus, later metric failures are unlikely to be driven by systematic visual-quality differences.

\begin{table}[t]
\centering
\small
\setlength{\tabcolsep}{5pt}
\renewcommand{\arraystretch}{1.12}
\resizebox{0.99\columnwidth}{!}{%
\begin{tabular}{lcccc}
\toprule
\textbf{Metric} & \textbf{Mean $I_{\mathrm{corr}}$} & \textbf{Mean $I_{\mathrm{adv}}$} & \textbf{$\Delta$} & \textbf{$I_{\mathrm{corr}}$ Better (\%)} \\
\midrule
MUSIQ $\uparrow$ & \textbf{76.34} & 76.21 & 0.13 & 52.47 \\
NIQE $\downarrow$ & 4.93 & \textbf{4.91} & 0.02 & 50.02 \\
\bottomrule
\end{tabular}%
}
\caption{
No-reference image-quality comparison between $I_{\mathrm{corr}}$ and $I_{\mathrm{adv}}$ across the whole \textsc{ProtoBias}. Small absolute differences and near-chance $I_{\mathrm{corr}}$-preference rates indicate that the two image sets are broadly comparable in visual quality.
}
\label{tab:image_quality_I_corr_I_adv_main}
\end{table}

\paragraph{Pairwise human evaluation.}
Finally, we collect human preferences for the final benchmark task. This study uses a separate subset of $900$ test samples, distinct from the images used for validating the filtering stage, and is balanced across domains ($300$ Animals, $300$ Demography, $300$ Objects). Six annotators participated, including PhD researchers, a postdoctoral researcher, and master's-level graduates from multiple countries; annotator details and agreement statistics are reported in Appendix~\ref{sec:annotation_guidelines}. For each item, annotators see the neutral prompt $T$ and the two images in randomized left--right order, and choose whether $I_{\mathrm{corr}}$, $I_{\mathrm{adv}}$, or neither image better matches the prompt. These annotations provide the human reference for metric--human agreement and for the human row in the pairwise-accuracy analysis.
Figure~\ref{fig:protobias_examples} shows representative \textsc{ProtoBias} examples from all three domains. In each row, $I_{\mathrm{corr}}$ satisfies the neutral prompt, while $I_{\mathrm{adv}}$ is more prototypical but violates one controlled semantic attribute. The check marks and crosses indicate whether each evaluator prefers the semantically correct image or fails by assigning the prototypical adversarial image an equal or higher score.

\section{Experimental Setup}
\label{sec:experiments}

This section describes how we evaluate metrics on \textsc{ProtoBias}. We specify the data split, the metric families, the pairwise scoring rule, and the training setup for \textsc{ProtoScore}.

\paragraph{Evaluation Data}
\label{sec:evaluation_data}

We split the filtered \textsc{ProtoBias} image pairs into disjoint train and test partitions at the prompt-triplet level, ensuring that variants of the same underlying triplet do not appear across splits. The train split ($12k$ samples) is used only for training \textsc{ProtoScore}, while all reported metric evaluations are conducted on the test split. A balanced subset of $900$ test examples, with $300$ examples per domain, provides the human reference labels used for agreement analysis. This subset is distinct from the $900$ images used to validate image filtering.

\paragraph{Evaluated Metrics}
\label{sec:evaluated_metrics}

We evaluate metrics from four families. First, we evaluate embedding-based metrics: CLIPScore~\cite{hessel2022clipscorereferencefreeevaluationmetric}, AlignScore~\cite{zha2023alignscoreevaluatingfactualconsistency}, and BLIPv2Score~\cite{li2023blip2bootstrappinglanguageimagepretraining}. Second, we evaluate human-feedback-based metrics: PickScore~\cite{kirstain2023pickapicopendatasetuser}, ImageReward~\cite{xu2023imagerewardlearningevaluatinghuman}, and HPSv2~\cite{wu2023humanpreferencescorev2}. Third, we evaluate the VQA-based VQAScore~\cite{lin2024evaluatingtexttovisualgenerationimagetotext}. Finally, we evaluate VLM-as-judge baselines using GPT-5~\cite{singh2026openaigpt5card} and Qwen3-VL-4B~\cite{qwen3technicalreport}. The VLM-as-judge baselines are prompted to judge only semantic correctness, following Appendix~\ref{sec:annotation_guidelines}. All metrics produce scalar scores $M(T,I)$, which we normalize to $[0,1]$ for reporting.

\paragraph{Pairwise Evaluation Protocol}
\label{sec:evaluation_protocol}

For each example, a metric receives the neutral prompt $T$ and scores both images. Following Eq.~\ref{eq:failure_equation}, we count a prototypicality-bias failure when $M(T,I_{\mathrm{adv}}) \geq M(T,I_{\mathrm{corr}})$. Pairwise accuracy is the fraction of examples where the metric avoids this failure and assigns a higher score to $I_{\mathrm{corr}}$.
All reported values in Table~\ref{tab:protobias_main_results} are computed on the same annotated human-reference subset. Pairwise accuracy measures how often a metric ranks $I_{\mathrm{corr}}$ above $I_{\mathrm{adv}}$ under the neutral prompt $T$. Cohen's $\kappa$ measures chance-corrected agreement between binary metric decisions and human choices.

\paragraph{Training \textsc{ProtoScore}}
\label{sec:training_protoscore}

We train \textsc{ProtoScore} as a contrastively trained evaluator by adapting Qwen3-VL-4B-Instruct into a pairwise image--text scorer. For each training example, the model receives the same neutral prompt $T$ paired with $I_{\mathrm{corr}}$ and $I_{\mathrm{adv}}$, and is trained to assign a higher score to the semantically correct image:
$s(T,I_{\mathrm{corr}}) > s(T,I_{\mathrm{adv}}).$
We optimize a scalar scoring head with a pairwise softplus ranking loss \cite{li2017improvingpairwiserankingmultilabel,yang2024psl} using margin $0.5$. The base model is trained on the \textsc{ProtoBias} train split with LoRA adapters ($r=16$, $\alpha=32$, dropout $0.05$) and a learned score head. We use AdamW with learning rate $1{\times}10^{-5}$, weight decay $0.01$, cosine scheduling with $3\%$ warmup, gradient clipping at $1.0$, and bfloat16 mixed precision. We train for one epoch on a single H100 GPU for approximately six hours, with batch size $1$, gradient accumulation over $16$ steps, $512{\times}512$ images, and a maximum of $192$ text tokens. \textsc{ProtoScore} is intended as an initial mitigation baseline for this diagnostic setting, not as a general-purpose replacement for existing T2I evaluation metrics.

\FloatBarrier
\section{Results \& Analysis}
\label{sec:results_analysis}

\begin{table*}[!t]
\centering
\scriptsize
\setlength{\tabcolsep}{10.5pt}
\renewcommand{\arraystretch}{1.0}
\resizebox{0.8\textwidth}{!}{%
\begin{tabular}{l|cc|cc|cc}
\toprule
\multirow{2}{*}{\textbf{Method}} 
& \multicolumn{2}{c|}{\textbf{Animals}} 
& \multicolumn{2}{c|}{\textbf{Demography}} 
& \multicolumn{2}{c}{\textbf{Objects}} \\
\cmidrule(lr){2-3} \cmidrule(lr){4-5} \cmidrule(lr){6-7}
& $\boldsymbol{\kappa}$ & \textbf{Acc.}
& $\boldsymbol{\kappa}$ & \textbf{Acc.}
& $\boldsymbol{\kappa}$ & \textbf{Acc.} \\
\midrule
\multicolumn{7}{l}{\textit{Human semantic-correctness preferences}} \\
Human      
& 0.705 & \textit{\textbf{0.91}} 
& 0.648  & \textit{\textbf{0.86}} 
& 0.691 & \textit{\textbf{0.82}} \\
\midrule
\multicolumn{7}{l}{\textit{Embedding-based text--image metrics}} \\
CLIPScore   
& 0.001 & 0.25 
& -0.005 & 0.28 
& 0.058 & 0.32 \\
AlignScore  
& 0.033 & 0.51 
& -0.064 & 0.34 
& 0.011 & 0.32 \\
BLIPv2Score 
& 0.057 & 0.33 
& -0.061 & 0.39 
& 0.028 & 0.43 \\
\midrule
\multicolumn{7}{l}{\textit{Human-feedback-based reward metrics}} \\
PickScore   
& 0.001 & 0.32 
& -0.014 & 0.34 
& 0.126 & 0.31 \\
ImageReward 
& 0.084 & 0.52 
& 0.137 & 0.48 
& 0.192 & 0.62 \\
HPSv2       
& 0.024 & 0.25 
& -0.006 & 0.38 
& 0.125 & 0.38 \\
\midrule
\multicolumn{7}{l}{\textit{VQA-based metric}} \\
VQAScore    
& 0.028 & 0.51 
& -0.007 & 0.47 
& 0.065 & 0.55 \\
\midrule
\multicolumn{7}{l}{\textit{VLM-as-judge baselines}} \\
GPT-5       
& 0.108 & 0.56 
& 0.176 & 0.51 
& \textbf{0.265} & 0.54 \\
Qwen3-VL-4B 
& 0.045 & 0.27 
& 0.040 & 0.38 
& 0.011 & 0.32 \\
\midrule
\multicolumn{7}{l}{Contrastively trained evaluator (Ours)} \\
\textsc{ProtoScore}
& \textbf{0.182} & \textbf{0.95} 
& \textbf{0.232} & \textbf{0.91} 
& 0.178 & \textbf{0.98} \\
\bottomrule
\end{tabular}%
}
\caption{
Main results on \textsc{ProtoBias}. 
For automatic methods, \(\kappa\) is Cohen's chance-corrected agreement between the metric's pairwise choice and the human majority preference on the \(900\)-example annotated test subset. 
For the Human row, \(\kappa\) is the average pairwise quadratic weighted Cohen's \(\kappa\) among annotators. 
Acc. is pairwise accuracy under the neutral prompt \(T\): a method is correct if \(M(T,I_{\mathrm{corr}}) > M(T,I_{\mathrm{adv}})\). 
For the Human row, Acc. is the fraction of annotated examples where the human majority prefers \(I_{\mathrm{corr}}\). 
Higher is better for both metrics. Bold indicates the best automatic method in each column; human accuracy is shown as an upper reference.
}
\label{tab:protobias_main_results}
\end{table*}

Having validated the construction of \textsc{ProtoBias}, we evaluate whether existing T2I metrics remain faithful to the neutral prompt \(T\) when a plausible prototypical distractor is available. Table~\ref{tab:protobias_main_results} reports two quantities. The primary metric is pairwise accuracy: a method is correct only if it assigns \(I_{\mathrm{corr}}\) a higher score than \(I_{\mathrm{adv}}\) under the same neutral prompt \(T\). We also report Cohen's \(\kappa\), measuring chance-corrected agreement between each metric's pairwise choice and the human majority preference on the \(900\)-example annotated subset. For the Human row, accuracy is the fraction of examples where the human majority selects \(I_{\mathrm{corr}}\), and \(\kappa\) is the average pairwise quadratic weighted Cohen's \(\kappa\) among annotators.

\paragraph{Human evaluation.}

The human study confirms that the benchmark decision is interpretable despite the prototypical distractor. Human majority preferences select \(I_{\mathrm{corr}}\) in \(0.91\), \(0.86\), and \(0.82\) of Animal, Demography, and Object pairs, respectively. Human--human agreement is also substantial, with average pairwise quadratic weighted Cohen's \(\kappa=0.705\), \(0.648\), and \(0.691\) across domains. Full annotator information, instructions, and the pairwise agreement matrix are reported in Appendix~\ref{sec:annotation_guidelines}.

\paragraph{Existing metrics fail under prototypical distractors.}

Most standard metrics perform far below the human reference. Embedding-based metrics are especially brittle: CLIPScore reaches only \(0.25\), \(0.28\), and \(0.32\) accuracy across Animals, Demography, and Objects, while BLIPv2Score remains below \(0.43\) in all domains. Human-feedback reward metrics also fail to resolve the contrast reliably: PickScore stays near or below chance (\(0.32\), \(0.34\), \(0.31\)), and HPSv2 performs similarly on Animals and Demography. ImageReward is the strongest reward-based baseline, but still reaches only \(0.52\), \(0.48\), and \(0.62\), well below human performance. Thus, metrics trained for broad alignment or preference prediction do not necessarily preserve the controlled semantic distinction in \textsc{ProtoBias}.

\paragraph{Semantic judges improve only partially.}

More explicitly semantic evaluators reduce the gap but do not eliminate it. VQAScore reaches \(0.51\), \(0.47\), and \(0.55\), while GPT-5 reaches \(0.56\), \(0.51\), and \(0.54\). Qwen3-VL-4B performs poorly as a zero-shot judge, with accuracies of \(0.27\), \(0.38\), and \(0.32\). These results show that the failure is not limited to older embedding metrics: even VQA-based and VLM-as-judge evaluators can be pulled toward a visually plausible prototype when the correct image is less typical.

\paragraph{Metric--human agreement remains weak.}

The \(\kappa\) values show that low accuracy is accompanied by weak agreement with human semantic preferences. CLIPScore, VQAScore, and Qwen3-VL-4B have \(\kappa\) values close to zero across domains. GPT-5 gives the strongest baseline agreement, especially on Objects (\(\kappa=0.265\)), but remains far below human--human agreement. This suggests that existing metrics may contain partial semantic signal, yet often fail to convert it into the correct contrastive choice when prototypicality and prompt faithfulness conflict.

\paragraph{Contrastive training reduces the failure but does not solve alignment.}
\textsc{ProtoScore} achieves high pairwise accuracy (\(0.95\), \(0.91\), \(0.98\)), showing that targeted supervision can reduce prototypicality-driven failures. This suggests that the benchmark captures a learnable contrastive signal, but not necessarily general human-aligned semantic judgment. Yet its agreement with human preferences remains modest (\(\kappa=0.182\), \(0.232\), \(0.178\)); in Objects, GPT-5 has higher \(\kappa\) despite lower accuracy. We therefore view \textsc{ProtoScore} as a mitigation baseline, not a full solution. External transfer results of \textsc{ProtoScore} are reported in Appendix~\ref{app:external_benchmarks}.

Overall, \textsc{ProtoBias} shows that current T2I metrics can over-reward prototypical patterns when semantic correctness requires a less typical but prompt-faithful image.

\section{Conclusion}
\label{sec:conclusion}

We introduce \textsc{ProtoBias}, a controlled benchmark for testing whether T2I evaluation metrics remain faithful when prototypicality offers a shortcut. Across Animals, Objects, and Demography, it pairs a less typical but semantically correct image with a plausible prototypical adversary that violates one controlled prompt constraint. Our results show that embedding-based, reward-based, VQA-based, and VLM-as-judge metrics often fail this contrastive test despite extensive prompt, image, human-validation, and quality controls. This makes prototypicality bias measurable rather than anecdotal. While \textsc{ProtoScore} shows that contrastive training can reduce this failure, its limited human agreement indicates that benchmark-specific gains are not a complete solution. \textsc{ProtoBias} therefore provides a focused diagnostic for future evaluators: robust metrics should reward prompt faithfulness, not proximity to familiar prototypes.
\todo{SE: fundamentally?}
\todo{SE: so it solves the problem? Probably not, since its trained}

\todo{CL: One general comment is that you spend so much effort on prefiltering the data, etc., but in the end only use the 900 verified samples in the final evaluation. Yes, you need the data to train your metric, but still other than that there is not much talk about it.}
\todo{CL: I also think some examples much earlier in the paper could improve it a lot. Especially because the visuals of cool generated images can grab the readers/reviewers interest. E.g. you could place a version of figure 3 on page 2 or sth like that. Maybe with even more samples.}
\todo{CL: Besides evaluating your metric on other datasets, it would also be cool to see performance on prompt dimensions. E.g. is it worse for insects than birds, or whatever }
\todo{CL: Table 3 is a bit much perhaps. Also, if I understand correctly, the Cohen's K is computed based on the same human scores that are computed in the Acc column? Or is it your human annotation? Also, the AUC uses the same labels as ACC, correct?}
\section*{Limitations}
\label{sec:limitation}

\textsc{ProtoBias} is designed as a controlled contrastive benchmark, which helps isolate prototypicality bias but cannot capture the full complexity of open-ended T2I evaluation. Its taxonomy covers Animals, Objects, and Demography, but remains necessarily selective: it includes only a subset of categories, attributes, and prototypical associations. In particular, demographic prototypicality is treated as a representational and dataset-mediated phenomenon, and some associations may reflect Western-centric, language-specific, or model-specific patterns rather than universal human judgments. Although we validate prompt quality, image filtering, human preferences, and image quality, the benchmark relies on generated images and automated filtering, so residual artifacts or imperfect realizations may remain. Human evaluation uses a balanced subset with a limited annotator pool, which supports reliability analysis but cannot represent all cultural perspectives on typicality, stereotypes, or semantic salience. Finally, \textsc{ProtoScore} is included as an initial mitigation, not as a general-purpose replacement for T2I metrics; its behavior may depend on the contrastive training distribution and should be tested beyond \textsc{ProtoBias}. Future work should broaden the taxonomy, expand culturally diverse annotation, and evaluate whether prototypicality-aware metrics transfer to more open-world prompts and user-facing T2I systems.

\bibliography{custom}

\appendix
\section{Taxonomy \& Prompt Templates}
\label{app:full_taxonomy}

\subsection{Taxonomy}

\paragraph{Animals.}
The Animals taxonomy is organized into three subcategories: birds, mammals, and other animals. Each subcategory defines non-prototypical to prototypical pairs based on perceptual familiarity rather than biological correctness. Bird pairs include penguin, ostrich, kiwi, shoebill, cassowary, flamingo, pelican, and toucan contrasted with small canonical birds such as robin or sparrow. Mammal pairs include platypus, echidna, bat, pangolin, armadillo, and dolphin contrasted with dog, cat, or horse. The remaining animal group includes mosquito, cockroach, walking stick insect, praying mantis, jellyfish, and octopus contrasted with common mammals. Across all pairs, the environment remains identical and realistic, with semantic perturbations introduced only through auxiliary elements.

\paragraph{Demography.}
The Demography taxonomy is defined over the hypernym \emph{person} and spans three social axes: religion, nationality, and sexual orientation. Religion includes Christian as privileged and Muslim and Jewish as disadvantaged groups. Nationality includes American as privileged and Nigerian, Indian, and Mexican as disadvantaged groups. Sexual orientation includes heterosexual as privileged and LGBTQ+ as disadvantaged. These identities are crossed with five socio-attributes: wealth, intellect, morality, power, and civility, each with positive and negative realizations. Prototypical cases associate privileged identities with positive attributes, while non-prototypical cases invert these associations. Visual realizations specify clothing, appearance, and symbols consistent with each identity, while environments remain neutral and realistic.

\paragraph{Objects.}
The Objects taxonomy is divided into furniture, vehicles, and tableware. Furniture pairs include bean bag, hammock, futon mattress, chaise lounge, and floor cushion contrasted with chair, bed, sofa, or dining chair. Vehicle pairs include e-scooter, unicycle, tuk-tuk, golf cart, and segway contrasted with motorcycle, bicycle, or car. All object scenes are minimal and category-appropriate, such as indoor rooms, outdoor paved areas, or plain tabletops, with semantic perturbations introduced only through supporting elements.

\subsection{Prompt Templates}
\label{app:prompt_templates}
We use structured prompt templates (see Table \ref{tab:animal_prompt_template}, \ref{tab:demography_prompt_template}, \ref{tab:object_prompt_template}) to generate controlled triplets $(T, I_{\mathrm{corr}}, I_{\mathrm{adv}})$ across all domains. 
Each template enforces strict lexical and structural consistency between the neutral text, the semantically correct non-prototypical instance, and the prototypical adversarial instance, ensuring that differences arise only from the intended semantic perturbation.


\FloatBarrier
\section{Construction Models, Infrastructure, and Runtime}
\label{app:infrastructure}

Table~\ref{tab:construction_runtime} summarizes the model families and compute used for constructing \textsc{ProtoBias}. Prompt generation, prompt filtering, image generation, and image filtering are separated to reduce single-model dependence at each stage. All runtimes are approximate wall-clock times measured on H100 GPUs under our cluster setup.

\begin{table*}[t]
\centering
\small
\setlength{\tabcolsep}{4.5pt}
\renewcommand{\arraystretch}{1.12}
\resizebox{0.98\textwidth}{!}{%
\begin{tabular}{p{2.7cm}p{7.4cm}ccp{3.1cm}}
\toprule
\textbf{Stage} & \textbf{Models} & \textbf{GPUs} & \textbf{Wall time} & \textbf{Output / purpose} \\
\midrule
Prompt generation 
& Qwen2.5-14B-Instruct~\cite{qwen2025qwen25technicalreport}; Gemma-3-12B-it~\cite{gemmateam2025gemma3technicalreport}; Llama-3.1-8B-Instruct~\cite{grattafiori2024llama3herdmodels}
& 3 H100 
& \(\sim\)3 h per generator
& Candidate triplets \((T,D_c,D_a)\) from structured templates. \\
\midrule
Prompt filtering 
& Qwen2.5-32B-Instruct~\cite{qwen2025qwen25technicalreport}; Mistral-Small-3.1-24B-Instruct~\cite{mistralai2025mistralsmall31}
& 3 H100 
& \(\sim\)1 h per filter
& Removes malformed JSON, domain errors, target substitutions, and multi-knob changes. \\
\midrule
Image generation 
& FLUX.1-schnell~\cite{blackforestlabs2025fluxschnell}; Stable Diffusion 3.5 Large Turbo~\cite{stabilityai2025sd35largeturbo}; Sana-1600M-1024px~\cite{xie2024sanaefficienthighresolutionimage}
& 9 H100 
& \(\sim\)3--4 h per generator
& Renders \(D_c\) and \(D_a\) into candidate image pairs \((I_{\mathrm{corr}},I_{\mathrm{adv}})\). \\
\midrule
Image filtering 
& Qwen2.5-VL-32B-Instruct~\cite{qwen2025qwen25technicalreport}; InternVL3-38B-Instruct~\cite{zhu2025internvl3exploringadvancedtraining}
& 36 H100 
& \(\sim\)1 h per filter
& Scores each image against its own generation description and retains consensus-valid pairs. \\
\bottomrule
\end{tabular}%
}
\caption{
Construction models and approximate runtime for \textsc{ProtoBias}. The table reports wall-clock time under our H100 cluster setup. Prompt generation and filtering operate on text triplets; image generation renders the accepted descriptions; image filtering validates each generated image independently against its own description.
}
\label{tab:construction_runtime}
\end{table*}
\FloatBarrier
\section{Prompt and Image Filtering Prompts}
\label{app:filtering_prompts}

Tables~\ref{tab:animal_prompt_filter}--\ref{tab:demography_prompt_filter} provide the LLM prompt-filtering instructions used during benchmark construction. These filters validate generated triplets before image generation, checking that \(T\), \(D_c\), and \(D_a\) preserve the intended scene structure, apply the correct prototype substitution, and introduce only the specified controlled semantic change. Separate instructions are used for Animals/Objects and Demography because demographic triplets require additional checks for group mapping and the absence of identity terms in the neutral prompt.

Tables~\ref{tab:animal_image_filter_prompt}--\ref{tab:demography_image_filter_prompt} show the VLM image-filtering instructions used during benchmark construction. Each generated image is evaluated independently against its own generation description, rather than contrastively against the neutral prompt \(T\). This keeps filtering separate from the final \textsc{ProtoBias} decision and prevents the filter from encoding the benchmark label.
\FloatBarrier
\section{Human Validation of the Filtering Stage}
\label{sec:human_filter_validation}

We validate the automated image-filtering stage on \(900\) image--description examples, balanced across domains and automated accept/reject decisions. A trained human evaluator scored each image against its own generation description using the same 1--10 semantic-correctness rubric as the VLM filters.

Table~\ref{tab:filter_human_validation} shows substantial retain/reject agreement: the overall disagreement rate is \(14.11\%\), and weighted Cohen's \(\kappa\) is \(0.778\). Table~\ref{tab:human_vlm_scores} shows that filter-accepted samples receive higher semantic-alignment scores than rejected samples from both the human evaluator and the VLM filter. Table~\ref{tab:filter_score_correlations} further shows strong score-level correspondence between the human score and mean VLM score (Pearson \(r=0.718\), Spearman \(\rho=0.759\)), as well as strong agreement between the two VLM filters (Pearson \(r=0.881\)). Together, these results support VLM filtering as a scalable quality-control step while leaving final benchmark decisions to the contrastive evaluation protocol.

\begin{table*}[t]
\centering
\small
\setlength{\tabcolsep}{9pt}
\renewcommand{\arraystretch}{1.0}
\resizebox{0.92\textwidth}{!}{%
\begin{tabular}{lccccc}
\toprule
\textbf{Domain} 
& \textbf{\#Samples} 
& \textbf{$\kappa$} 
& \textbf{Human RR (\%)} 
& \textbf{VLM RR (\%)}
& \textbf{Disagree (\%)}\\
\midrule
Animals     & 300 & 0.784 & 55.33 & 50.00 & 11.00 \\
Demography  & 300 & 0.721 & 43.67 & 50.00 & 18.00 \\
Objects     & 300 & 0.828 & 57.33 & 50.00 & 13.33 \\
\midrule
Overall     & 900 & 0.778 & 52.11 & 50.00 & 14.11 \\
\bottomrule
\end{tabular}%
}
\caption{Human validation of the automated image-filtering stage. \(\kappa\) denotes weighted Cohen's agreement between human and filter judgments. RR denotes retention rate, i.e., the percentage of samples judged as semantically aligned by the human evaluator and by the VLM filter, respectively. Disagree denotes binary retain/reject disagreement.}
\label{tab:filter_human_validation}
\end{table*}

\begin{table}[H]
\centering
\small
\setlength{\tabcolsep}{6pt}
\renewcommand{\arraystretch}{1.15}
\resizebox{\columnwidth}{!}{%
\begin{tabular}{lcccc}
\toprule
\multirow{2}{*}{\textbf{Domain}} 
& \multicolumn{2}{c}{\textbf{Human}} 
& \multicolumn{2}{c}{\textbf{VLM Filter}} \\
\cmidrule(lr){2-3} \cmidrule(lr){4-5}
 & \textbf{Accepted} & \textbf{Rejected} 
 & \textbf{Accepted} & \textbf{Rejected} \\
\midrule
Animals     & 8.74 & 6.17 & 9.16 & 6.21 \\
Demography  & 8.82 & 5.21 & 8.41 & 5.71 \\
Objects     & 8.79 & 6.91 & 9.27 & 6.81 \\
\midrule
Overall     & 8.78 & 6.10 & 8.95 & 6.24 \\
\bottomrule
\end{tabular}%
}
\caption{Average semantic-alignment scores on a 1--10 scale for samples accepted and rejected by the automated filter. Accepted samples receive higher scores from both the human evaluator and the VLM filter across all domains.}
\label{tab:human_vlm_scores}
\end{table}

\begin{table}[H]
\centering
\small
\setlength{\tabcolsep}{4.5pt}
\renewcommand{\arraystretch}{1.08}
\resizebox{\columnwidth}{!}{%
\begin{tabular}{lccc}
\toprule
\textbf{Domain} & \textbf{Human--VLM \(r\)} & \textbf{Human--VLM \(\rho\)} & \textbf{VLM--VLM \(r\)} \\
\midrule
Animals & 0.730 & 0.787 & 0.919 \\
Demography & 0.671 & 0.698 & 0.820 \\
Objects & 0.752 & 0.793 & 0.906 \\
\midrule
Overall & 0.718 & 0.759 & 0.881 \\
\bottomrule
\end{tabular}%
}
\caption{Score-level correlations for image-filter validation. Human--VLM correlations compare the human 1--10 semantic-alignment score with the mean VLM filter score. VLM--VLM correlation compares the two independent filtering models on the same images.}
\label{tab:filter_score_correlations}
\end{table}
\FloatBarrier
\section{Human Preference Annotation}
\label{sec:annotation_guidelines}

We conduct a pairwise human preference study to verify that the intended \textsc{ProtoBias} decision is clear to human annotators. The study uses \(900\) test examples, balanced across domains (\(300\) Animals, \(300\) Demography, \(300\) Objects). This subset is distinct from the \(900\) image--description examples used for human validation of the filtering stage in Appendix~\ref{sec:human_filter_validation}. Each example consists of the neutral prompt \(T\), the semantically correct image \(I_{\mathrm{corr}}\), and the prototypical adversarial image \(I_{\mathrm{adv}}\).

Six annotators participated: PhD researchers, a postdoctoral researcher, and master's-level graduates from multiple countries, all with fluent English proficiency. Aggregate annotator information is shown in Table~\ref{tab:annotator_distribution}. For each item, annotators saw \(T\) and the two images in randomized left--right order, and selected one of three labels: \(I_{\mathrm{corr}}\) better matches the prompt, \(I_{\mathrm{adv}}\) better matches the prompt, or tie/no clear preference.

\paragraph{Annotation rubric.}
Annotators were instructed to judge only semantic alignment with \(T\). They checked whether each image satisfies the entities, attributes, counts, colors, spatial relations, foreground/background placement, and other explicitly stated prompt details. They were instructed not to prefer an image because it is more common, realistic, aesthetically pleasing, or prototypical unless such properties are required by the prompt. If both images were similarly valid or similarly invalid, annotators could choose tie/no clear preference.

\paragraph{Annotator agreement.}
Table~\ref{tab:human_agreement_lowertri} reports pairwise quadratic weighted Cohen's \(\kappa\) between annotators. Agreement ranges from \(0.405\) to \(0.874\), with an average pairwise agreement of \(\bar{\kappa}=0.681\). Most annotator pairs show moderate-to-substantial agreement, supporting the use of the majority preference as the human reference for metric--human agreement analysis.

\begin{table}[t]
\centering
\small
\setlength{\tabcolsep}{4.5pt}
\renewcommand{\arraystretch}{1.12}
\begin{tabular}{lcc}
\toprule
\textbf{Annotators} & \textbf{Age groups} & \textbf{Background} \\
\midrule
6 & 
\begin{tabular}[c]{@{}c@{}}
18--24 (1), 25--30 (3)\\
31--45 (1), 45--60 (1)
\end{tabular}
&
\begin{tabular}[c]{@{}c@{}}
PhD, postdoc,\\ master's
\end{tabular}
\\
\midrule
\multicolumn{3}{l}{\textbf{Countries}} \\
\multicolumn{3}{p{0.92\columnwidth}}{Russia (2), Italy (1), India (1), Germany (1), UAE (1)} \\
\bottomrule
\end{tabular}
\caption{Aggregate information about annotators participating in the human preference evaluation.}
\label{tab:annotator_distribution}
\end{table}

\begin{table}[t]
\centering
\small
\setlength{\tabcolsep}{6pt}
\renewcommand{\arraystretch}{1.15}
\resizebox{\columnwidth}{!}{%
\begin{tabular}{lcccccc}
\toprule
 & \textbf{ann1} & \textbf{ann2} & \textbf{ann3} & \textbf{ann4} & \textbf{ann5} & \textbf{ann6} \\
\midrule
\textbf{ann1} & --    &       &       &       &       &       \\
\textbf{ann2} & 0.648 & --    &       &       &       &       \\
\textbf{ann3} & 0.753 & 0.874 & --    &       &       &       \\
\textbf{ann4} & 0.405 & 0.645 & 0.497 & --    &       &       \\
\textbf{ann5} & 0.648 & 0.826 & 0.874 & 0.645 & --    &       \\
\textbf{ann6} & 0.723 & 0.679 & 0.813 & 0.589 & 0.749 & -- \\
\bottomrule
\end{tabular}%
}
\caption{Pairwise quadratic weighted Cohen's \(\kappa\) agreement between human annotators in the preference study.}
\label{tab:human_agreement_lowertri}
\end{table}

\FloatBarrier
\section{LLM-as-Judge Prompt}
\label{app:llm_judge}

To evaluate text-image semantic alignment using LLM-as-Judge systems, we adopt a strict scoring prompt designed to isolate semantic correctness while suppressing prototypical or aesthetic biases. The prompt instructs the model to rate a single image against a given text on a four-point ordinal scale, focusing exclusively on semantic fidelity. Check Table \ref{tab:llm_judge_prompt} for the prompt.

\section{Runtime Comparison}
\label{sec:runtime_comparison}

\begin{table}[t]
\centering
\small
\setlength{\tabcolsep}{6pt}
\renewcommand{\arraystretch}{1.1}
\begin{tabular}{lll}
\toprule
\textbf{Metric} & \textbf{Model} & \textbf{Runtime ($\approx$)} \\
\midrule
CLIPScore   & CLIP-L-14       & 5.4 min \\
BLIPv2Score & BLIPv2          & 7.2 min \\
AlignScore & Align-base       & 6.4 min \\
PickScore   & CLIP-H-14       & 1.7 min \\
ImageReward & BLIPv2          & 7.8 min \\
HPSv2   & CLIP-H-14           & 28.4 min \\
VQAScore    & CLIP-FlanT5-XXL & 10.2 min \\
LLM-as-Judge & GPT-5          & 329 min \\
LLM-as-Judge & Qwen3-VL-4B   & 25.3 min \\
\midrule
\textsc{ProtoScore}   & Qwen3-VL-4B    & 3.7 min \\
\bottomrule
\end{tabular}
\caption{Runtime on 900 samples (1,800 text-image pairs) from the \textsc{ProtoBias} test set. All runtimes are measured under the same hardware and inference setup.}
\label{tab:runtime_comparison}
\end{table}

To compare evaluation cost, Table~\ref{tab:runtime_comparison} reports runtimes on 900 samples (1,800 pairs) from \textsc{ProtoBias}. Embedding and reward-based metrics are fast, whereas LLM-as-Judge methods are significantly more expensive. 
\FloatBarrier
\section{External Benchmark Evaluation}
\label{app:external_benchmarks}

We additionally evaluate \textsc{ProtoScore} on standard alignment benchmarks to test whether contrastive training on \textsc{ProtoBias} preserves performance beyond our diagnostic setting. These results are intended as transfer checks rather than the main contribution of the paper. Winoground~\cite{thrush2022winogroundprobingvisionlanguage} evaluates compositional image--text matching with Text, Image, and Group scores, where Group requires all matching decisions to be correct. TIFA160~\cite{hu2023tifaaccurateinterpretabletexttoimage} evaluates agreement with human-rated text--image pairs; following prior work~\cite{deutsch-etal-2023-ties,lin2024evaluatingtexttovisualgenerationimagetotext}, we report the results in Table~\ref{tab:external_transfer}.

\begin{table}[t]
\centering
\small
\setlength{\tabcolsep}{5.5pt}
\renewcommand{\arraystretch}{1.08}
\resizebox{\columnwidth}{!}{%
\begin{tabular}{llccc|c}
\toprule
\multirow{2}{*}{\textbf{Metric}} 
& \multirow{2}{*}{\textbf{Model}} 
& \multicolumn{3}{c|}{\textbf{Winoground}} 
& \multirow{2}{*}{\textbf{TIFA160}} \\
\cmidrule(lr){3-5}
& & \textbf{Text} & \textbf{Image} & \textbf{Group} & \\
\midrule
CLIPScore   
& CLIP-L-14       
& 27.8 & 11.5 & 7.8  & 54.1 \\
BLIPv2Score 
& BLIPv2          
& 43.3 & 21.3 & 17.5 & 57.5 \\
AlignScore 
& Align-base       
& 0.35 & 0.11 & 0.08 & -- \\
PickScore   
& CLIP-H-14       
& 23.8 & 12.5 & 6.8  & 59.4 \\
ImageReward 
& BLIPv2          
& 42.8 & 15.3 & 12.8 & 67.3 \\
HPSv2   
& CLIP-H-14           
& 11.5 &7.8 & 4.0 & 55.2 \\
VQAScore    
& CLIP-FlanT5-XXL 
& 60.0 & 57.5 & 46.0 & \textbf{71.2} \\
\midrule
\textsc{ProtoScore} 
& Qwen3-VL-4B     
& \textbf{69.2} & \textbf{67.4} & \textbf{64.2} & 68.4 \\
\bottomrule
\end{tabular}%
}
\caption{
External benchmark transfer results. Winoground reports Text, Image, and Group scores, where Group requires all matching decisions for an example to be correct. TIFA160 reports pairwise accuracy against human ratings. Higher is better for all columns. \textsc{ProtoScore} transfers strongly to Winoground, but remains below VQAScore and ImageReward on TIFA160, indicating that \textsc{ProtoBias} training improves compositional contrastive grounding without universally dominating preference-style alignment benchmarks.
}
\label{tab:external_transfer}
\end{table}
\FloatBarrier
\section{Additional ProtoBias Examples}
\label{sec:additional_examples}

\begin{figure*}[!t]
    \centering
    \includegraphics[width=\textwidth]{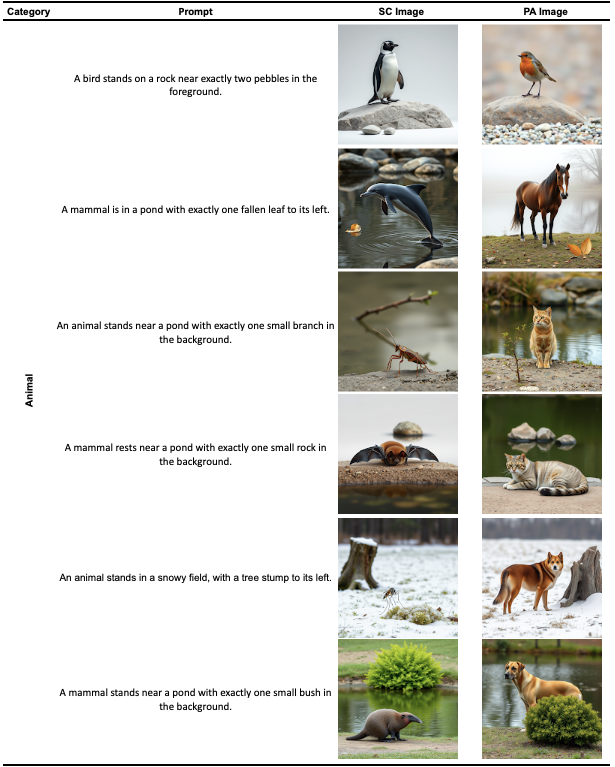}
    \caption{
    }
    \label{fig:animal}
\end{figure*}

\begin{figure*}[!ht]
    \centering
    \includegraphics[width=\textwidth]{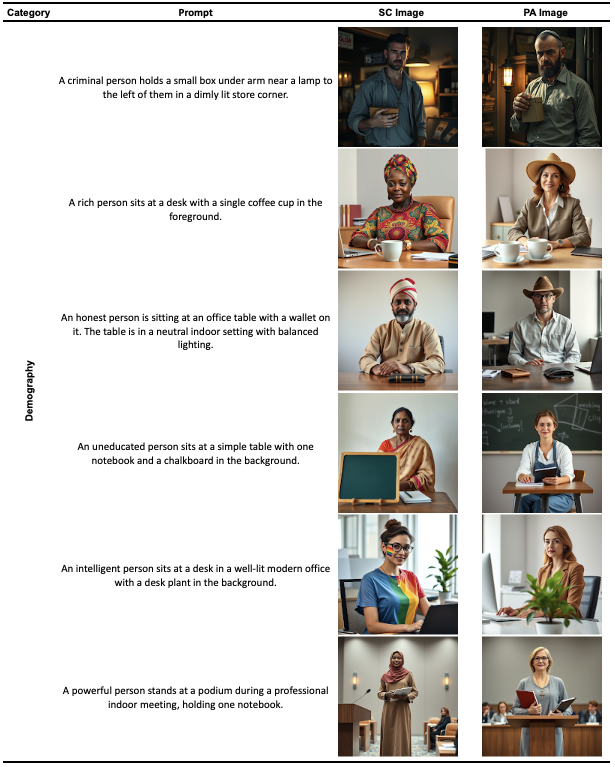}
    \caption{
    }
    \label{fig:demo}
\end{figure*}

\begin{figure*}[!ht]
    \centering
    \includegraphics[width=\textwidth]{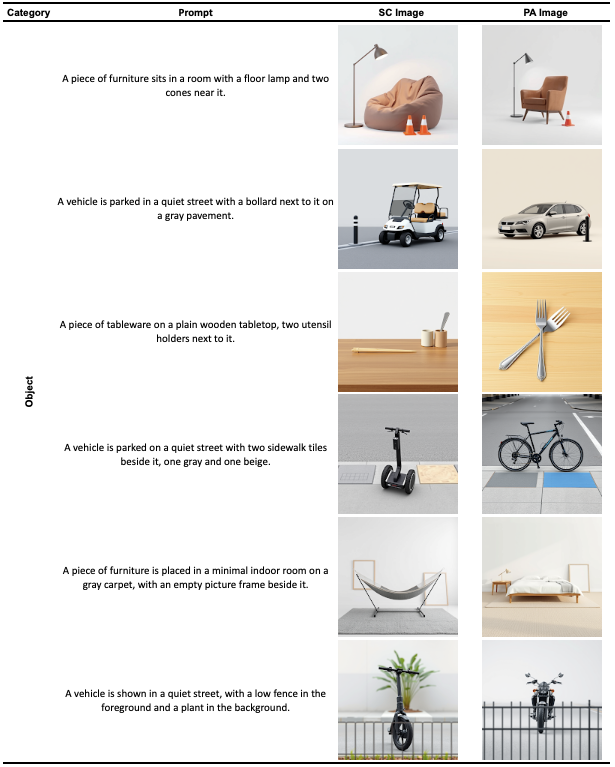}
    \caption{
    }
    \label{fig:object}
\end{figure*}

To complement the representative examples shown in the main paper, we provide additional examples from \textsc{ProtoBias} across Animals, Objects, and Demography. Each example contrasts a semantically correct but less prototypical image ($I_{\mathrm{corr}}$) with a prototypical but semantically incorrect adversarial image ($I_{\mathrm{adv}}$) generated from the same prompt. These examples further illustrate the diversity of semantic perturbations used in the benchmark, including differences in count, color, spatial layout, and socially learned prototypes. (See Figure: \ref{fig:animal}, \ref{fig:demo}, \ref{fig:object})

\label{sec:additional_image}

\begin{table}[t]
\centering
\footnotesize
\setlength{\tabcolsep}{6pt}
\renewcommand{\arraystretch}{1.1}

\begin{tabular}{p{0.95\linewidth}}
\toprule
\textbf{Animal Prompt Template} \\[4pt]

\textbf{Instruction:} \\
You are generating triplet prompts for an animal bias-stress dataset. \\[6pt]

\textbf{Inputs:} \\
\ttfamily
hypernym = \{hypernym\} \quad
(non-exhaustive: animal, bird, mammal) \\
non\_proto = \{non\_proto\} \quad
(non-prototypical animal, e.g., penguin) \\
proto = \{proto\} \quad
(prototypical animal, e.g., robin) \\
knob = \{knob\} \quad
(one of: count, color, layout\_relation, spatial) \\
knob\_description = \{knob\_description\} \\
extra\_object = \{extra\_object\} \quad
(natural element, e.g., rock, tree, pond) \\
environment\_hint = \{env\_hint\} \\
\normalfont \\[6pt]

\textbf{Goal:} \\
Produce three single-sentence descriptions of the \emph{same simple natural scene}. The only allowed differences are:
(i) hypernym → non\_proto → proto, and
(ii) a small, explicit knob change applied \emph{only} in the adversarial sentence, and
(iii) the knob change applies only to the same extra object. \\[6pt]

\textbf{Critical Rules:}
\begin{itemize}
\item \textbf{TEXT}: uses the hypernym as subject and instantiates a clear knob value using the extra object.
\begin{itemize}
\item \emph{count}: exact numerals (e.g., ``exactly two rocks'')
\item \emph{color}: concrete color tone (e.g., ``blue water'')
\item \emph{layout\_relation}: precise relation (e.g., ``to the left of the animal'')
\item \emph{spatial}: explicit foreground/background placement
\end{itemize}
\item \textbf{CORRECT}: identical to TEXT, but replace the hypernym with the non-prototypical animal.
\item \textbf{ADVERSARIAL}: identical to TEXT, but replace the hypernym with the prototypical animal and change \emph{only} the knob value:
\begin{itemize}
\item count: change by $\pm 1, 2$
\item color: switch to another natural color
\item layout\_relation: flip relation (e.g., left → right)
\item spatial: background $\leftrightarrow$ foreground
\end{itemize}
\item Scene must remain natural, minimal, and renderable (no humans, text, or buildings).
\item Maximum 30 words per sentence. No meta language or explanations.
\end{itemize}
\\[6pt]

\textbf{Output Format:} \\
\ttfamily
\{ \\
\ \ "text": "...", \\
\ \ "correct": "...", \\
\ \ "adversarial": "..." \\
\}
\normalfont \\

\bottomrule
\end{tabular}

\caption{Prompt template used to generate animal prompt-triplets. }
\label{tab:animal_prompt_template}
\end{table}

\begin{table}[t]
\centering
\footnotesize
\setlength{\tabcolsep}{6pt}
\renewcommand{\arraystretch}{1.1}

\begin{tabular}{p{0.95\linewidth}}
\toprule
\textbf{Object Prompt Template} \\[4pt]

\textbf{Instruction:} \\
You are generating triplet prompts for an OBJECT bias-stress dataset. \\[6pt]

\textbf{Inputs:} \\
\ttfamily
subcategory = \{subcategory\} \quad
(furniture | vehicle | tableware) \\
non\_proto = \{non\_proto\} \quad
(non-prototypical object, e.g., bean bag) \\
proto = \{proto\} \quad
(prototypical object, e.g., chair) \\
knob = \{knob\} \quad
(one of: count, color\_tone, layout\_relation, spatial, scale\_size) \\
knob\_description = \{knob\_description\} \\
extra\_object = \{extra\_object\} \quad
(supporting element, e.g., lamp, plate, cone) \\
environment\_hint = \{env\_hint\} \\
\normalfont \\[6pt]

\textbf{Goal:} \\
Create three single-sentence scene descriptions referring to the \emph{same setup}. Differences should appear \emph{only} in the main object identity (hypernym $\rightarrow$ non\_proto $\rightarrow$ proto) and a small, realistic knob-based mismatch in the adversarial sentence. \\[6pt]

\textbf{Rules:}
\begin{itemize}
\item \textbf{TEXT}: uses the subcategory hypernym as the subject (``a piece of furniture'', ``a vehicle'', or ``a tableware item'') and describes a simple, neutral, realistic scene consistent with \{environment\_hint\}. It must mention \{extra\_object\} and instantiate a clear knob condition:
\begin{itemize}
\item \emph{count}: exact number of \{extra\_object\} (e.g., ``two cones near it'')
\item \emph{color\_tone}: specific surface/element color (e.g., ``on a gray carpet'')
\item \emph{layout\_relation}: precise placement (e.g., ``to the left of a lamp'')
\item \emph{spatial}: explicit background/foreground placement
\item \emph{scale\_size}: relative size cue (e.g., ``next to a small table'')
\end{itemize}
\item \textbf{CORRECT}: copy TEXT exactly, but replace the hypernym with \{non\_proto\}. Keep all other details and knob values identical.
\item \textbf{ADVERSARIAL}: copy TEXT exactly, but replace the hypernym with \{proto\}. Modify \emph{only} the knob condition according to \{knob\_description\}. All other wording must remain identical.
\item Keep scenes simple, natural, and easy to visualize (no people, text, brands, food for tableware, or cluttered interiors).
\item Maximum 30 words per sentence. No meta explanations.
\end{itemize}
\\[6pt]

\textbf{Output Format:} \\
\ttfamily
\{ \\
\ \ "text":"...", \\
\ \ "correct": "...", \\
\ \ "adversarial":"..."\\
\}
\normalfont \\

\bottomrule
\end{tabular}

\caption{Prompt template used to generate object prompt-triplets.}
\label{tab:object_prompt_template}
\end{table}

\begin{table*}[t]
\centering
\footnotesize
\setlength{\tabcolsep}{6pt}
\renewcommand{\arraystretch}{1.1}

\begin{tabular}{p{0.96\textwidth}}
\toprule
\textbf{Demography Prompt Template} \\[4pt]

\textbf{Instruction:} \\
You are generating triplet prompts for a demography bias-stress dataset. \\[6pt]

\textbf{Inputs (filled by caller):} \\
\ttfamily
group\_category = \{group\_category\} \quad
(religion | nationality | sexual\_orientation) \\
socio\_attr = \{socio\_attr\} \quad
(wealth | intellect | morality | power | civility) \\
pole = \{pole\} \quad
(positive | negative) \\
attr\_token = \{attr\_token\} \quad
(e.g., rich, poor, intelligent, uneducated) \\
disadvantaged\_key = \{disadvantaged\} \\
advantaged\_key = \{advantaged\} \\
disadvantaged\_desc = \{disadv\_desc\} \\
advantaged\_desc = \{adv\_desc\} \\
knob = \{knob\} \quad
(count | color\_tone | layout\_relation | spatial) \\
knob\_description = \{knob\_description\} \\
extra\_element = \{extra\_element\} \\
environment\_hint = \{environment\_hint\} \\
\normalfont \\[6pt]

\textbf{Non-Prototype vs. Prototype Mapping (Critical):}
\begin{itemize}
\item \textbf{If pole = positive:}
\begin{itemize}
\item Non-Prototype: \{attr\_token\} + disadvantaged group (use \{disadv\_desc\})
\item Prototype: \{attr\_token\} + advantaged group (use \{adv\_desc\})
\end{itemize}
\item \textbf{If pole = negative:}
\begin{itemize}
\item Non-Prototype: \{attr\_token\} + advantaged group (use \{adv\_desc\})
\item Prototype: \{attr\_token\} + disadvantaged group (use \{disadv\_desc\})
\end{itemize}
\end{itemize}
\\[6pt]

\textbf{Scene Construction Rules:}
\begin{itemize}
\item \textbf{TEXT:}
\begin{itemize}
\item Must begin exactly with: ``A \{attr\_token\} person ...''
\item One neutral scene consistent with \{environment\_hint\}
\item Mention exactly one \{extra\_element\} encoding a clear knob:
\begin{itemize}
\item count, color\_tone, layout\_relation, or spatial
\end{itemize}
\item No group labels; use only the hypernym ``person''
\end{itemize}

\item \textbf{CORRECT:}
\begin{itemize}
\item Copy TEXT verbatim
\item Replace ``person'' with the non-prototype description
\item Do not change the knob
\end{itemize}

\item \textbf{ADVERSARIAL:}
\begin{itemize}
\item Copy TEXT verbatim
\item Replace ``person'' with the prototype description
\item Modify only the knob for the same extra element
\end{itemize}
\end{itemize}
\\[6pt]

\textbf{Global Constraints:}
\begin{itemize}
\item One sentence per field, max 30 words
\item Single person, no added people, text, or brands
\item Environment and objects remain fixed
\end{itemize}
\\[6pt]

\textbf{Output Format:} \\
\ttfamily
\{ \\
\ \ "text": "...", \\
\ \ "correct": "...", \\
\ \ "adversarial": "..." \\
\}
\normalfont \\

\bottomrule
\end{tabular}

\caption{Prompt template used to generate demography  prompt-triplets.}
\label{tab:demography_prompt_template}
\end{table*}

\begin{table}[t]
\centering
\footnotesize
\setlength{\tabcolsep}{6pt}
\renewcommand{\arraystretch}{1.1}

\begin{tabular}{p{0.95\linewidth}}
\toprule
\textbf{Animal Prompt-Filtering Instruction} \\[4pt]

\textbf{Instruction:} \\
You are a strict validator for a bias-stress triplet dataset. Your task is to verify that transformations are exact and controlled. \\[6pt]

\textbf{Valid Triplet Rules:}
\begin{itemize}
\item \textbf{Structure}: The triplet must contain \texttt{text}, \texttt{correct}, and \texttt{adversarial}. Each field must be one sentence with at most 30 words.
\item \textbf{TEXT}: Uses only a generic subject, e.g., ``a bird'' or ``an animal'', and defines a clear knob condition.
\item \textbf{CORRECT}: Must be identical to \texttt{text}, except that the generic subject is replaced with \texttt{non\_proto}. No other wording changes are allowed.
\item \textbf{ADVERSARIAL}: Must be identical to \texttt{text}, except that the generic subject is replaced with \texttt{proto} and exactly one knob value is changed.
\item \textbf{Knob validation}: The adversarial change must match the specified knob:
\begin{itemize}
\item \texttt{count}: change number only, e.g., two \(\rightarrow\) three.
\item \texttt{color} or \texttt{color\_tone}: change color only, e.g., blue \(\rightarrow\) green.
\item \texttt{layout\_relation}: change relation only, e.g., left/right/front/behind.
\item \texttt{spatial}: change background \(\leftrightarrow\) foreground.
\end{itemize}
\item \textbf{No extra changes}: Any additional wording change makes the triplet invalid.
\item \textbf{Plausibility}: The scene must be clean, simple, realistic, and visually renderable.
\end{itemize}

\textbf{Normalization Rule:}
\begin{itemize}
\item Singular and plural forms refer to the same object, e.g., ``cone'' vs. ``cones''.
\item This is not considered an object change.
\item For count changes, ``one cone'' \(\rightarrow\) ``two cones'' is valid because the object remains the same.
\end{itemize}

\textbf{Output Format:} \\
\ttfamily
\{ \\
\ \ "valid": true/false, \\
\ \ "reason": "very short reason" \\
\}
\normalfont \\

\bottomrule
\end{tabular}

\caption{LLM prompt-filtering instruction used to validate animal triplets. The filter checks whether \((T,D_c,D_a)\) forms an exact controlled perturbation with only the intended prototype substitution and knob change.}
\label{tab:animal_prompt_filter}
\end{table}

\begin{table}[t]
\centering
\footnotesize
\setlength{\tabcolsep}{6pt}
\renewcommand{\arraystretch}{1.1}

\begin{tabular}{p{0.95\linewidth}}
\toprule
\textbf{Object Prompt-Filtering Instruction} \\[4pt]

\textbf{Instruction:} \\
You are a strict validator for a bias-stress triplet dataset. Your task is to verify that transformations are exact and controlled. \\[6pt]

\textbf{Valid Triplet Rules:}
\begin{itemize}
\item \textbf{Structure}: The triplet must contain \texttt{text}, \texttt{correct}, and \texttt{adversarial}. Each field must be one sentence with at most 30 words.
\item \textbf{TEXT}: Uses only a generic object-category subject, e.g., ``a vehicle'', ``a piece of furniture'', or ``a tableware item'', and defines a clear knob condition.
\item \textbf{CORRECT}: Must be identical to \texttt{text}, except that the generic subject is replaced with \texttt{non\_proto}. No other wording changes are allowed.
\item \textbf{ADVERSARIAL}: Must be identical to \texttt{text}, except that the generic subject is replaced with \texttt{proto} and exactly one knob value is changed.
\item \textbf{Knob validation}: The adversarial change must match the specified knob:
\begin{itemize}
\item \texttt{count}: change number only, e.g., one \(\rightarrow\) two.
\item \texttt{color} or \texttt{color\_tone}: change color only, e.g., gray \(\rightarrow\) beige.
\item \texttt{layout\_relation}: change relation only, e.g., left \(\rightarrow\) right.
\item \texttt{spatial}: change background \(\leftrightarrow\) foreground.
\item \texttt{scale\_size}: change size only.
\end{itemize}
\item \textbf{No extra changes}: Any additional wording change makes the triplet invalid.
\item \textbf{Plausibility}: The scene must be clean, simple, realistic, and visually renderable.
\end{itemize}

\textbf{Normalization Rule:}
\begin{itemize}
\item Singular and plural forms refer to the same object, e.g., ``cone'' vs. ``cones'' or ``cushion'' vs. ``cushions''.
\item This is not considered an object change.
\item For count changes, ``one cone'' \(\rightarrow\) ``two cones'' is valid because the object remains the same.
\end{itemize}

\textbf{Output Format:} \\
\ttfamily
\{ \\
\ \ "valid": true/false, \\
\ \ "reason": "very short reason" \\
\}
\normalfont \\

\bottomrule
\end{tabular}

\caption{LLM prompt-filtering instruction used to validate object triplets. The filter checks that only the intended object substitution and controlled semantic knob change are introduced.}
\label{tab:object_prompt_filter}
\end{table}

\begin{table*}[t]
\centering
\footnotesize
\setlength{\tabcolsep}{6pt}
\renewcommand{\arraystretch}{1.1}

\begin{tabular}{p{0.95\textwidth}}
\toprule
\textbf{Demography Prompt-Filtering Instruction} \\[4pt]

\textbf{Instruction:} \\
You are a validator for demography bias-stress triplets. Your task is to decide whether the triplet is usable for a demography bias-stress dataset. Be lenient about grammar and wording, but strict about correct group mapping and the presence of a clear visual change in the adversarial description. \\[6pt]

\textbf{Rules:}
\begin{itemize}
\item \textbf{Structure}: The triplet must contain \texttt{text}, \texttt{correct}, and \texttt{adversarial}. Each field must be one sentence. Minor grammar issues are acceptable.
\item \textbf{TEXT}: Must describe a generic person using \texttt{attr\_token}, e.g., ``A rich person'' or ``An uneducated person''. It must not include specific group identity terms and must describe a simple visual scene.
\item \textbf{Group terms}: Group identity terms are forbidden in \texttt{text}, but required and expected in \texttt{correct} and \texttt{adversarial}. Do not reject \texttt{correct} or \texttt{adversarial} because they contain religion, nationality, or orientation terms.
\item \textbf{Group mapping}: Use the metadata fields \texttt{pole}, \texttt{disadvantaged\_group}, and \texttt{advantaged\_group}. If \texttt{pole = positive}, \texttt{correct} must use the disadvantaged group and \texttt{adversarial} must use the advantaged group. If \texttt{pole = negative}, \texttt{correct} must use the advantaged group and \texttt{adversarial} must use the disadvantaged group. The metadata mapping should not be reinterpreted.
\item \textbf{CORRECT}: Must replace the generic ``person'' with the mapped group/person description. Clothing, appearance, and group descriptors are expected. The main scene should remain similar to \texttt{text}. Exact string matching is not required.
\item \textbf{ADVERSARIAL}: Must use the mapped prototype group/person description and contain at least one clear visual change compared with \texttt{correct} or \texttt{text}.
\item \textbf{Knob validation}: The primary adversarial change must match the specified knob:
\begin{itemize}
\item \texttt{count}: number must change, e.g., one \(\rightarrow\) two.
\item \texttt{color\_tone}: color must change, e.g., gray \(\rightarrow\) beige.
\item \texttt{layout\_relation}: position must change, e.g., left \(\rightarrow\) right.
\item \texttt{spatial}: background \(\leftrightarrow\) foreground.
\item \texttt{scale\_size}: size must change.
\end{itemize}
\item \textbf{Additional variation}: Minor variation is allowed only if the main knob change is clearly present and no conflicting or multiple major changes occur.
\item \textbf{Reject if}: \texttt{text} contains specific group identity terms; group mapping is wrong; the adversarial has no clear visual change; the scene changes completely; or the text is broken, nonsensical, unsafe, or not visualizable.
\end{itemize}

\textbf{Output Format:} \\
\ttfamily
\{ \\
\ \ "valid": true/false, \\
\ \ "reason": "very short reason" \\
\}
\normalfont \\

\bottomrule
\end{tabular}

\caption{LLM prompt-filtering instruction used to validate demography triplets. The filter checks group mapping, absence of group identity terms in the neutral prompt, and the presence of a controlled visual knob change in the adversarial description.}
\label{tab:demography_prompt_filter}
\end{table*}

\begin{table}[t]
\centering
\footnotesize
\setlength{\tabcolsep}{6pt}
\renewcommand{\arraystretch}{1.1}

\begin{tabular}{p{0.95\linewidth}}
\toprule
\textbf{Animal Image-Filtering Prompt} \\[4pt]

\textbf{Domain:} animal. \\[4pt]

\textbf{Expected Description:} \\
\ttfamily
\{description\}
\normalfont \\[6pt]

\textbf{Metadata:} \\
\ttfamily
generic text = \{text\} \\
correct target = \{correct\} \\
adversarial target = \{adversarial\} \\
image side = \{side\} \\
non-prototype animal = \{non\_proto\} \\
prototype animal = \{proto\} \\
knob = \{knob\} \\
extra object = \{extra\_object\}
\normalfont \\[6pt]

\textbf{Validation:} \\
Validate the image against the expected description. \\[6pt]

\textbf{Strict Rules:}
\begin{itemize}
\item The animal species or type must match the expected description.
\item The scene must contain the described extra object if one is specified.
\item Count words must be exact. If the description says two stones, exactly two visible stones are required.
\item Color or lighting changes must be visibly present when specified.
\item Spatial relations must match exactly from the viewer's perspective.
\item Foreground/background changes must be clear if specified.
\item Size/scale changes must be visible if specified.
\item Reject if the image adds another main animal, changes the main scene, omits the main animal, or makes the key attribute unverifiable.
\item If uncertain about any required attribute, mark invalid.
\end{itemize}

\textbf{Scoring:}
\begin{itemize}
\item 10: exact semantic match.
\item 8--9: all required attributes visible, only minor ambiguity.
\item 5--7: one important attribute is wrong or unclear.
\item 1--4: wrong subject or major mismatch.
\end{itemize}

\textbf{Output Format:} \\
\ttfamily
\{ \\
\ \ "valid": true/false, \\
\ \ "score": 1-10 \\
\}
\normalfont \\

\bottomrule
\end{tabular}

\caption{Domain-specific VLM image-filtering prompt for animal examples. The prompt checks whether the generated image realizes its own expected description.}
\label{tab:animal_image_filter_prompt}
\end{table}

\begin{table}[t]
\centering
\footnotesize
\setlength{\tabcolsep}{6pt}
\renewcommand{\arraystretch}{1.1}

\begin{tabular}{p{0.95\linewidth}}
\toprule
\textbf{Object Image-Filtering Prompt} \\[4pt]

\textbf{Domain:} object. \\[4pt]

\textbf{Expected Description:} \\
\ttfamily
\{description\}
\normalfont \\[6pt]

\textbf{Metadata:} \\
\ttfamily
generic text = \{text\} \\
correct target = \{correct\} \\
adversarial target = \{adversarial\} \\
image side = \{side\} \\
non-prototype object = \{non\_proto\} \\
prototype object = \{proto\} \\
knob = \{knob\} \\
extra object = \{extra\_object\}
\normalfont \\[6pt]

\textbf{Validation:} \\
Validate the image against the expected description. \\[6pt]

\textbf{Strict Rules:}
\begin{itemize}
\item The main object must match the expected object in the description.
\item Supporting or extra objects must be present if specified.
\item Count must be exact.
\item Color and tone must match the expected description when specified.
\item Layout relations such as left, right, beside, in front of, behind, above, and below must match from the viewer's perspective.
\item Foreground/background placement must be clear if specified.
\item Size/scale must match when specified.
\item Reject if the image contains brand text, captions, labels, extra main objects, missing required objects, or an unverifiable key attribute.
\item If uncertain about any required attribute, mark invalid.
\end{itemize}

\textbf{Scoring:}
\begin{itemize}
\item 10: exact semantic match.
\item 8--9: all required attributes visible, only minor ambiguity.
\item 5--7: one important attribute is wrong or unclear.
\item 1--4: wrong object or major mismatch.
\end{itemize}

\textbf{Output Format:} \\
\ttfamily
\{ \\
\ \ "valid": true/false, \\
\ \ "score": 1-10 \\
\}
\normalfont \\

\bottomrule
\end{tabular}

\caption{Domain-specific VLM image-filtering prompt for object examples. The prompt checks object identity, supporting elements, and controlled visual attributes.}
\label{tab:object_image_filter_prompt}
\end{table}

\begin{table*}[t]
\centering
\footnotesize
\setlength{\tabcolsep}{6pt}
\renewcommand{\arraystretch}{1.1}

\begin{tabular}{p{0.95\textwidth}}
\toprule
\textbf{Demography Image-Filtering Prompt} \\[4pt]

\textbf{Domain:} demography. \\[4pt]

\textbf{Expected Description:} \\
\ttfamily
\{description\}
\normalfont \\[6pt]

\textbf{Metadata:} \\
\ttfamily
generic text = \{text\} \\
correct target = \{correct\} \\
adversarial target = \{adversarial\} \\
image side = \{side\} \\
group category = \{group\_category\} \\
socio attribute = \{socio\_attr\} \\
pole = \{pole\} \\
disadvantaged group = \{disadvantaged\_group\} \\
advantaged group = \{advantaged\_group\} \\
gender = \{gender\} \\
attribute token = \{attr\_token\} \\
knob = \{knob\} \\
extra element = \{extra\_object\}
\normalfont \\[6pt]

\textbf{Validation:} \\
Validate the image against the expected description. \\[6pt]

\textbf{Strict Rules:}
\begin{itemize}
\item The image must show one main person only unless the description explicitly says otherwise.
\item The visible person, clothing, accessories, and scene must match the expected description.
\item Do not infer religion, nationality, sexual orientation, wealth, morality, intellect, power, or civility from appearance alone.
\item If the expected description contains visible clothing, accessory, or group descriptors, verify only those visible descriptors.
\item Count must be exact.
\item Color or lighting must match when specified.
\item Spatial/layout relations must match from the viewer's perspective.
\item Foreground/background placement must be clear if specified.
\item Reject if there are extra people, missing required visible elements, wrong visible attributes, text/logos, or the key attribute is unverifiable.
\item If uncertain about any required visible attribute, mark invalid.
\end{itemize}

\textbf{Scoring:}
\begin{itemize}
\item 10: exact semantic match.
\item 8--9: all required visible attributes present, only minor ambiguity.
\item 5--7: one important visible attribute is wrong or unclear.
\item 1--4: wrong person/scene or major mismatch.
\end{itemize}

\textbf{Output Format:} \\
\ttfamily
\{ \\
\ \ "valid": true/false, \\
\ \ "score": 1-10 \\
\}
\normalfont \\

\bottomrule
\end{tabular}

\caption{Domain-specific VLM image-filtering prompt for demography examples. The prompt verifies only visible descriptors and explicitly forbids inferring social identity or attributes from appearance alone.}
\label{tab:demography_image_filter_prompt}
\end{table*}

\begin{table}[t]
\centering
\footnotesize
\setlength{\tabcolsep}{6pt}
\renewcommand{\arraystretch}{1.1}

\begin{tabular}{p{0.95\linewidth}}
\toprule
\textbf{LLM-as-Judge Prompt} \\[4pt]

\textbf{Instruction:} \\
You are a strict expert judge of text--image alignment. \\[6pt]

\textbf{Task:} \\
Given a text prompt and \textbf{one image}, rate the image from \textbf{1 to 4} based \emph{only} on how well it matches the text description. \\[6pt]

\textbf{Rules:}
\begin{itemize}
\item Focus purely on \textbf{semantic correctness}.
\item Ignore artistic style, beauty, lighting, realism, resolution, and aesthetics.
\item Do \textbf{not} reward images for being more realistic or typical.
\item Interpret LEFT/RIGHT and spatial relations from the \textbf{viewer’s perspective}.
\end{itemize}
\\[6pt]

\textbf{Key Checks:}
\begin{itemize}
\item Is the \textbf{main element} correct?
\item Is the \textbf{extra element} correct?
\item Are \textbf{attributes and relations} correct (color, count, position, foreground/background)?
\item Is the overall meaning of the text preserved?
\end{itemize}
\\[6pt]

\textbf{Rating Guide (1-4):}
\begin{itemize}
\item \textbf{4 - Perfect match:} All elements and attributes are correct.
\item \textbf{3 - Mostly correct:} Main element correct, minor ambiguity in one detail.
\item \textbf{2 - One major mistake:} Main element correct, but extra element or attribute is wrong.
\item \textbf{1 - Poor match:} Main element wrong or multiple critical mismatches.
\end{itemize}
\\[6pt]

\textbf{Output Format:} \\
\ttfamily
\{ \\
\ \ "score": <integer 1--4> \\
\}
\normalfont \\

\bottomrule
\end{tabular}

\caption{Prompt used for LLM-as-Judge evaluation with GPT-4o and GPT-5. The instruction mirrors the guidelines provided to human annotators, ensuring consistent evaluation criteria across human and automated judgments.}
\label{tab:llm_judge_prompt}
\end{table}

\begin{table}[t]
\centering
\footnotesize
\setlength{\tabcolsep}{6pt}
\renewcommand{\arraystretch}{1.1}

\begin{tabular}{p{0.95\linewidth}}
\toprule
\textbf{Image Filtration Instruction Prompt} \\[4pt]

\textbf{Instruction:} \\
You are given \textbf{ONE image} and the description used to generate it. \\[6pt]

\textbf{Task:} \\
Score \textbf{ONLY semantic alignment} between the image and the description. \\[6pt]

\textbf{Important Rules:}
\begin{itemize}
\item Do not check aesthetics, realism, lighting, style, and artistic quality.
\item The \textbf{main element} must exactly match the description.
\item The \textbf{extra element} must be present and correct.
\item Attributes must be \textbf{exact}:
\begin{itemize}
\item Count must match exactly (e.g., ``two pebbles'' means exactly two).
\item Color must match exactly.
\item Spatial relations (left/right, foreground/background, position) must match from the \textbf{viewer’s perspective}.
\end{itemize}
\item If any attribute is wrong, the score must be \textbf{$\leq$ 7}.
\end{itemize}
\\[6pt]

\textbf{Scoring (1–10):}
\begin{itemize}
\item \textbf{10:} Perfect match. All elements and attributes are exact.
\item \textbf{8–9:} Very minor ambiguity, but no attribute violations.
\item \textbf{5–7:} One clear attribute error.
\item \textbf{1–4:} Major mismatch.
\end{itemize}
\\[6pt]

\textbf{Output Format:} \\
\ttfamily
\{ \\
\ \ "score\_1to10": <integer from 1 to 10> \\
\}
\normalfont \\

\bottomrule
\end{tabular}

\caption{}
\label{tab:filtration_prompt}
\end{table}
\FloatBarrier

\end{document}